\documentclass{article}

\usepackage{microtype}
\usepackage{graphicx}
\usepackage{booktabs} 

\usepackage{hyperref}

\usepackage{theorems}
\usepackage{shortcuts}
\usepackage{subcaption}
\usepackage[]{csquotes} 
\usepackage[inline]{enumitem}
\usepackage{tikz-qtree}
\usepackage{tikz}
\usetikzlibrary{bayesnet}

\usepackage{ulem}
\usepackage{adjustbox}
\usepackage{placeins}

\usepackage[accepted]{icml2021}
\icmltitlerunning{Regularizing towards Causal Invariance: Linear Models with Proxies}

\begin{document}

\twocolumn[
\icmltitle{Regularizing towards Causal Invariance: Linear Models with Proxies}

\icmlsetsymbol{equal}{*}

\begin{icmlauthorlist}
\icmlauthor{Michael Oberst}{mit} 
\icmlauthor{Nikolaj Thams}{cph}
\icmlauthor{Jonas Peters}{cph}
\icmlauthor{David Sontag}{mit}
\end{icmlauthorlist}

\icmlaffiliation{mit}{EECS, MIT, Cambridge, USA}
\icmlaffiliation{cph}{Department of Mathematical Sciences, University of Copenhagen, Copenhagen, Denmark}

\icmlcorrespondingauthor{Michael Oberst}{moberst@mit.edu}

\icmlkeywords{Machine Learning, Distributional Robustness, Causality, Anchor Regression, ICML}

\vskip 0.3in
]

\printAffiliationsAndNotice{}

\begin{abstract}
We propose a method for learning linear models whose predictive performance is robust to causal interventions on unobserved variables, when noisy proxies of those variables are available.  Our approach takes the form of a regularization term that trades off between in-distribution performance and robustness to interventions.  Under the assumption of a linear structural causal model, we show that a single proxy can be used to create estimators that are prediction optimal under interventions of bounded strength. This strength depends on the magnitude of the measurement noise in the proxy, which is, in general, not identifiable. In the case of two proxy variables, we propose a modified estimator that is prediction optimal under interventions up to a known strength.
We further show how to extend these estimators to scenarios where additional information about the \enquote{test time} intervention is available during training. We evaluate our theoretical findings in synthetic experiments and using real data of hourly pollution levels across several cities in China.
\end{abstract}

\section{Introduction}%
\label{sec:introduction}

Ideally, predictive models would generalize beyond the distribution on which they are trained, e.g., across geographic regions, across time, or across individual users.  However, 
models often learn to rely on signals in the training distribution that are not stable across domains, causing a drop-off in predictive performance. This problem is broadly known as dataset shift \citep{Quinonero-Candela2009}.

Tackling this problem requires a formalization of how dataset shift arises, and how that shift impacts the conditional distribution of our target $Y$ given features $X$.  One way to formalize this shift is in terms of an underlying causal graph \citep{Pearl2009}, where changes between distributions are seen as arising from causal interventions on variables.  

\begin{figure}[t]
  \centering
    \begin{tikzpicture}[
      obs/.style={circle, draw=gray!90, fill=gray!30, very thick, minimum size=5mm},
      uobs/.style={circle, draw=gray!90, fill=gray!10, dotted, minimum size=5mm},
      bend angle=30]
      \node[uobs] (A) {$A$} ;
      \node[obs] (Y) [below right=of A, yshift=6mm] {$Y$};
      \node[obs] (X) [below left=of A, yshift=6mm] {$X$} ;
      \draw[-latex, thick] (A) -- (X) node[midway, above left] {$\beta_{x}$};
      \draw[-latex, thick] (A) -- (Y) node[midway, above right] {$\beta_{y}$};
      \draw[-latex, thick] (X) -- (Y) node[midway, above] {$\alpha$};
    \end{tikzpicture}
    \caption{Conceptual Example: $A$ represents an (unobserved) socioeconomic variable, $X$ represents current health status, and $Y$ represents a long-term health outcome.  All relationships are assumed to be linear, and coefficients are given. We consider a broader class of graphs in this work, see Figure~\ref{fig:causal_graphs}.}%
    \label{fig:intro_simple_confounder}
\end{figure}
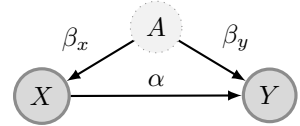

\textbf{Conceptual example}: In the causal graph given in Figure~\ref{fig:intro_simple_confounder}, the variable $A$ serves as a confounder.  In a medical setting, $A$ could represent smoking habits or socioeconomic status, which have a causal effect on current health status ($X$) as well as longer-term outcomes ($Y$).  Importantly, $A$ may not be recorded in our training data, and the distribution of $A$ could vary across geography and time.

In the context of this causal graph, interventions which change the distribution of $A$ will also alter the conditional mean $\E(Y \mid X)$. Under the linear relationships in Figure~\ref{fig:intro_simple_confounder}, the optimal least-squares predictor $\hat{Y} = \gamma^* X$ under the test distribution depends on the test-time variance in $A$, in that
\begin{align*}
  \gamma^* &= \begin{cases}
  \alpha, &\ \text{if after intervention } A = 0 \\
  \alpha + \frac{\beta_Y}{\beta_X}, &\ \text{if after intervention } \var(A) \rightarrow \infty. \\
  \end{cases}
\end{align*}
The first predictor encodes the direct causal effect of $X$ on $Y$, but is only optimal in the setting where the correlations induced by $A$ are removed by fixing it to a constant value of zero (the same holds when including intercepts and allowing for non-zero means).  The second predictor, on the other hand, renders the distribution of the residual $Y - \hat{Y}$ independent of $A$, and is therefore robust to arbitrary interventions upon $A$.  However, this is only optimal under arbitrarily strong interventions on $A$.

\textbf{Balancing performance and invariance}: Instead of seeking an invariant predictor that is robust to arbitrary interventions on $A$ (like the second predictor above), we instead seek to minimize a worst-case loss under bounded interventions of a given strength.
We contrast this with work that seeks to discover causal relationships as a route to invariance \citep{Rojas-Carulla2015, Magliacane2017}, optimize for invariance directly across environments \citep{Arjovsky2019-kv}, or use known causal structure to select predictors with invariant performance \citep{Subbaswamy2019}.

Our proposed objective takes the form of a standard loss, plus a regularization term that encourages invariance.  This builds upon \citet{Rothenhausler2018}, who introduce a similar objective, and prove that their objective optimizes a worst-case loss over bounded interventions on $A$, under a large class of linear structural causal models.

In contrast to \citet{Rothenhausler2018}, we do not assume that $A$ is observed.  Instead we assume that, during training, we have access to noisy proxies of $A$.  For most of the paper, we assume that neither $A$ nor proxies are available during testing.
With this in mind, our contributions are as follows
\begin{itemize}[itemsep=5pt,topsep=0pt,parsep=0pt,partopsep=0pt]
  \item \textit{Distributional robustness to bounded shifts}: In Section~\ref{sec:worst_case_bounded_shift}, we show that a single proxy can be used to construct estimators with distributional robustness guarantees under bounded interventions on $A$.  However, these estimators are robust to a strictly smaller set of interventions, compared to when $A$ is used directly, and the size of this set depends on the (unidentifiable) noise in the proxy.  When two proxies are available, we propose a modified estimator that can be used to recover the same guarantees as when $A$ is observed.
  \item \textit{Targeted shifts}: In Section~\ref{sec:targeted_shift}, we show how to target our loss to interventions on $A$ contained in a specified robustness set.  We show that this formulation includes Anchor Regression as a special case, but also allows for sets that are not centered around the mean of $A$.  In this setting we give an estimator, using two proxies, that identifies the target loss.
\end{itemize}
In Section~\ref{sec:experiments}, we evaluate our theoretical findings on synthetic experiments, and in Section~\ref{sec:experiment_pollution} we demonstrate our method on a real-world dataset consisting of hourly pollution readings across five major cities in China.

\section{Preliminaries}%
\label{sec:preliminaries}

\subsection{Notation}%
\label{sub:notation}

We use upper case letters $X$ to denote (possibly vector-valued) random variables, and lower-case letters $x$ to denote values in the range of those random variables. Vectors are assumed to be column vectors, so that $X \in \R^{d_X}$ indicates that $X = {(X_1, \ldots, X_{d_X})}^\top$, a column vector of $d_X$ random variables.  We use $\Sigma_X \in \R^{d_X \times d_X}$ to denote the covariance matrix of a variable $X$.  We use bold upper-case letters $\bX$ to denote a data matrix in $\R^{n \times d_X}$, consisting of $n$ i.i.d.\ observations of $X$, and $\1{\cdot}$ as an indicator random variable.  When dealing with matrices $C, D$, we use $C \prec D$ and $C \preceq D$ to indicate the positive definite and positive semi-definite partial order, respectively.  That is, $C \prec D$ if $D - C$ is positive definite (PD), and $C \preceq D$ if $D - C$ is positive semi-definite (PSD).  We use $\operatorname{Id}$ to denote the identity matrix, whose dimension is given by context.  All proofs are provided in the supplementary material.  

\subsection{Linear structural causal model}%
\label{sub:linear_structural_causal_model}

\begin{figure}[t]
  \centering
  \begin{tikzpicture}[
    obs/.style={circle, draw=gray!90, fill=gray!30, very thick, minimum size=7mm},
    uobs/.style={circle, draw=gray!90, fill=gray!10, dotted, minimum size=7mm},
    bend angle=10]
    \node[uobs] (H) {$H$} ;
    \node[uobs] (A) [left=of H, yshift=3mm] {$A$} ;
    \node[obs] (Y) [below right=of H, yshift=2mm, xshift=-7mm] {$Y$};
    \node[obs] (X) [below left=of H, yshift=2mm, xshift=7mm] {$X$} ;
    \node[obs] (W) [left=of X, xshift=5mm] {$W$} ;
    \node[obs] (Z) [left=of W, xshift=5mm] {$Z$} ;
    \draw[-latex, thick] (A) -- (X);
    \draw[-latex, thick] (A) -- (Y);
    \draw[-latex, thick] (A) -- (H);
    \draw[-latex, thick] (A) -- (W);
    \draw[-latex, thick] (A) -- (Z);
    \draw[-latex, thick] (H) to[bend left] (Y);
    \draw[-latex, thick] (Y) to[bend left] (H);
    \draw[-latex, thick] (H) to[bend left] (X);
    \draw[-latex, thick] (X) to[bend left] (H);
    \draw[-latex, thick] (Y) to[bend left] (X);
    \draw[-latex, thick] (X) to[bend left] (Y);
  \end{tikzpicture}
  \caption{In contrast to \citet{Rothenhausler2018}, we assume that anchor variables (denoted $A$) are unobserved, but that we have access to either one or two proxies $W, Z$.  
    Observed variables are shown in dark grey and unobserved variables in light grey.
    We do not assume knowledge of the causal structure between $A, X, H, Y$ (except that $A$ has no causal parents). The relationship between $X, H, Y$ could be cyclic, but all relationships are linear.
}%
\label{fig:causal_graphs}
\end{figure}
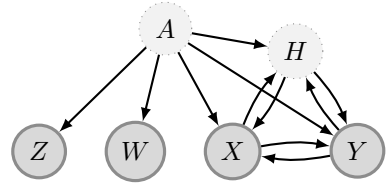

We assume the general class of causal graphs represented in Figure~\ref{fig:causal_graphs}, where $X \in \R^{d_X}$ denotes observed covariates that can be used in prediction, $Y \in \R^{d_Y}$ is the target we seek to predict, $H \in \R^{d_H}$ are unobserved variables, and $A \in \R^{d_A}$ represents anchor variables, which are assumed to have no causal parents in the graph.
We assume the linear structural causal model (SCM) given in Assumption~\ref{asmp:linear_scm}.
\begin{thmasmp}[Linear SCM]\label{asmp:linear_scm}
We assume the SCM 
\begin{equation}
  \begin{pmatrix} X \\ Y \\ H \end{pmatrix} \coloneqq B \begin{pmatrix} X \\ Y \\ H \end{pmatrix} + M_A A + \epsilon,
  \label{eq:ar_scm_full}
\end{equation}
where $A, \mathbf{\epsilon}$ have zero mean, bounded covariance, and are independently distributed. We assume that $\E[AA^\top]$ and $\operatorname{Id} - B$ are invertible, where $\operatorname{Id}$ is the identity matrix. See Figure~\ref{fig:causal_graphs} for a graphical representation.
\end{thmasmp}
Note that we do not assume here (or anywhere in this paper) that either $A$ or $\epsilon$ is Gaussian. The invertibility of $\operatorname{Id} - B$ is satisfied if the causal graph is a directed acyclic graph. The matrices $B, M_A$ encode the linear causal relationships.  For instance, Figure~\ref{fig:intro_simple_confounder} can be represented in this form by $B = \begin{bmatrix} 0 & 0 \\ \alpha & 0 \end{bmatrix}$, $M = \begin{bmatrix} \beta_X \\ \beta_Y \end{bmatrix}$.  In general, $\epsilon \in \R^{D}$, $B \in \R^{D \times D}$, and $M \in \R^{D \times d_A}$, where $D \coloneqq d_X + d_Y + d_H$.  We assume that $d_Y = 1$ for simplicity.

\subsection{Distributional robustness of anchor regression}\label{subsec:dist-robust-AR}

Our goal is to learn a predictor $f^*(X)$ of $Y$ that minimizes a worst-case risk of the following form
\begin{equation} 
  f^* = \argmin_{f \in \cF} \sup_{\P \in \cP} \E_{\P}[\ell(Y, f(X))],
  \label{eq:robust_objective}
\end{equation}
where $\cF$ denotes a hypothesis class of possible predictors, $\cP$ denotes a set of possible distributions, and $\ell$ represents our loss function.  We take the class $\cP$ to consist of distributions that arise as the result of causal interventions on $A$, and seek to learn a linear predictor to minimize mean-squared error.  

We use $\P$ to refer to the observational distribution, and $\P_{do(A \coloneqq \nu)}$ to refer to the distribution under interventions on $A$, where the variable $A$ is replaced by the random variable $\nu$, and $\nu$ is assumed to be independent of the noise vector $\epsilon$.  We often write \[R(\gamma) \coloneqq Y - \gamma^\top X\] as a random variable that represents the residual of a predictor $\gamma \in \R^{d_X}$.  Importantly, Assumption~\ref{asmp:linear_scm} implies that for any $\gamma$, $\E[R(\gamma) \mid A]$ can be written as a linear function in $A$.

In this setting, \citet{Rothenhausler2018} propose the following objective, defined here with respect to the observational distribution $\P$ (rather than a finite sample)
\begin{thmdef}[Anchor Regression]\label{def:ar_objective}
\begin{equation}
  \ell_{AR}(A; \gamma, \lambda) \coloneqq \ell_{LS}(X, Y; \gamma) + \lambda \ell_{PLS}(X, Y, A; \gamma), \label{eq:anchor_regression_objective}
\end{equation}
where $\lambda \geq -1$ is a hyperparameter and 
\begin{align}
  \ell_{LS}(X, Y; \gamma) &\coloneqq \E \left[ {R(\gamma)}^2 \right] \label{eq:def_ls} \\
  \ell_{PLS}(X, Y, A; \gamma) &\coloneqq \E\left[ {\left( \E \left[ R(\gamma) \mid A \right]\right)}^2 \right]\label{eq:def_pls}.
\end{align}
\end{thmdef}
The first term $\ell_{LS}$ encodes the least-squares objective, while the second term $\ell_{PLS}$ encodes the residual error which can be predicted from $A$, which we refer to as the projected least-squares error.  For $\lambda > 0$, the second term adds an additional penalty (beyond that of ordinary least squares) when the bias varies across values of $A$.
The second term~\eqref{eq:def_pls} can also be written in the linear setting of Assumption~\ref{asmp:linear_scm} as
\begin{equation}
  \ell_{PLS}(A; \gamma) = \E[R(\gamma) A^\top] {\E[AA^\top]}^{-1} \E[A {R(\gamma)}^{\top}]
  \label{eq:def_pls_linear},
\end{equation}
where we drop the dependence on $X, Y$ for notational simplicity.  Under Assumption~\ref{asmp:linear_scm}, Equation~\eqref{eq:anchor_regression_objective} corresponds to a worst-case loss under distributional shift caused by bounded intervention on $A$ \citep[][Theorem 1]{Rothenhausler2018}
\begin{align}\label{eq:arguarantee}
  \ell_{AR}(A; \gamma, \lambda) &= \sup_{\nu \in C_{A}(\lambda)} \E_{do(A \coloneqq \nu)}[{(Y - \gamma^\top X)}^2],
\end{align}
where the robustness set is given by 
\begin{equation}
  C_{A}(\lambda) \coloneqq \{ \nu: \E[\nu \nu^\top] \preceq (1 + \lambda) \E[AA^\top]\}.
  \label{eq:original_guarantee}
\end{equation}


Since minimizing $\ell_{AR}$ is equivalent to ordinary least squares (OLS) regression when $\lambda = 0$, this also provides a natural robustness guarantee for the OLS estimator, where $C_{OLS} \coloneqq \{ \nu: \E[\nu \nu^\top] \preceq \E[AA^\top] \}$.  In an identifiable instrumental variable setting, the minimizer converges against the causal parameter for $\lambda \rightarrow \infty$ \citep[e.g.][eq.\ (71)]{Jakobsen2020}; the $\ell_{PLS}$ term has therefore been referred to as `causal regularization' \citep[e.g.][]{Buhlmann2020b}, 
and has also been denoted by $\ell_{IV}$ \citep{Rothenhausler2018}, as $\cov(A, R(\gamma)) = \mathbf{0}$ if and only if $\ell_{PLS}(\gamma) = 0$.

\section{Distributional robustness to bounded shifts}%
\label{sec:worst_case_bounded_shift}

We first assume the existence of a noisy proxy $W$, conditionally independent of $(X, Y, H)$ given $A$ (see Figure~\ref{fig:causal_graphs}). 
\begin{thmasmp}[Single proxy with additive noise]\label{asmp:ar_scm_one_proxy}
In the context of Assumption~\ref{asmp:linear_scm}, $W$ is generated as follows
\begin{equation*}
  W \coloneqq  \beta_W^\top A + \epsilon_W,
\end{equation*}
where $\epsilon_W$ has mean zero, bounded covariance, and is independent of $(A, \epsilon)$.  In addition, we assume that the second moment matrix $\E[WW^\top]$ is invertible.
\end{thmasmp}
Under mild identifiability conditions (e.g., that $\beta_W$ is full rank) 
one can show (see Section~\ref{sub:add_proofs}) 
that 
\begin{equation}
  \ell_{PLS}(A; \gamma) = 0 \iff \ell_{PLS}(W; \gamma) = 0 \label{eq:cov_id},
\end{equation}
Hence, a single proxy is enough (in the population case) to identify whether the sharp constraint $\ell_{PLS}(\gamma) = 0$ holds, representing invariance to interventions of arbitrary strength.  This corresponds to the fact that if $A$ is a valid instrumental variable, then so is $W$ \citep{Hernan2006}.

However, we consider interventions on $A$ that are not of arbitrarily large strength.
With that in mind, in Section~\ref{sub:a_single_proxy_is_insufficient_for_robustness}, we demonstrate that 
\begin{enumerate*}[label=(\roman*)]
  \item when a single proxy $W$ is used in place of $A$, a robustness guarantee holds, but the robustness set is reduced relative to~\eqref{eq:original_guarantee}, 
  \item the extent of this reduction depends on the signal-to-variance relationship in $W$, and 
  \item this relationship is not generally identifiable from the observational distribution over $(X, Y, W)$ alone.  
\end{enumerate*}
In Section~\ref{sec:cross_estimator}, we show that in the setting where two proxies are available, the same guarantees as for an observed $A$ can be obtained.  We do so constructively, giving a regularization term whose population version is equal to $\ell_{PLS}(A; \gamma)$.

\subsection{Robustness with a single proxy}%
\label{sub:a_single_proxy_is_insufficient_for_robustness}
\begin{figure*}[t]
    \centering
  \input{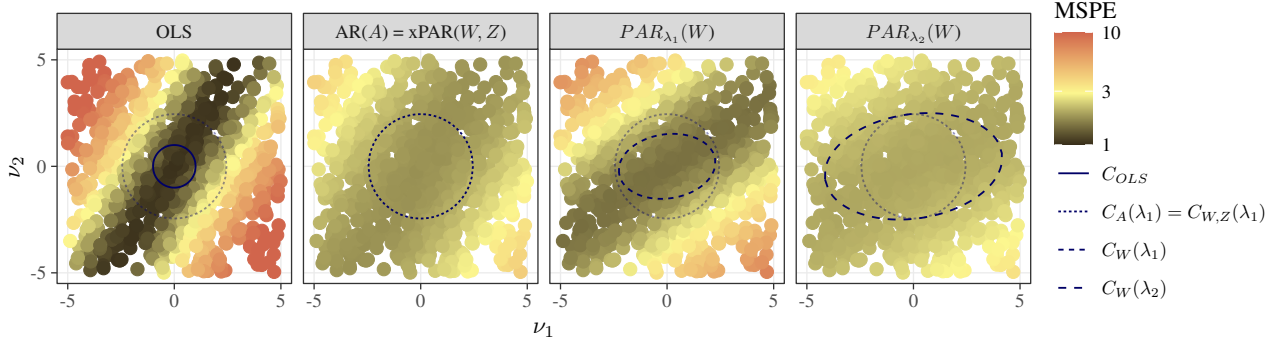}
\vspace{-10pt}
      \caption{Test performance under interventions $do(A \coloneqq (\nu_1, \nu_2))$ which give rise to different test distributions over $X$ and $Y$.
      Each dot corresponds to a different intervention (i.e., test distribution on $X, Y$), and the color gives the resulting mean squared prediction error (MSPE).
      \textbf{(Far Left)} OLS performs well for interventions in the set $C_{\text{OLS}}$ (solid circle), corresponding to the training covariance of $A$.  However, it performs poorly under interventions far from this region (e.g., top left).
      \textbf{(Middle Left)} Anchor Regression (AR) minimizes the worst-case loss over interventions on $A$ within the region $C_A(\lambda_1)$ (cf.,~\eqref{eq:original_guarantee}), a re-scaling of $C_{\text{OLS}}$.  There is a trade-off, with better performance than OLS under large interventions, but worse performance under small interventions. Given two proxies $W, Z$, we introduce Cross-Proxy Anchor Regression (xPAR, cf.,~\eqref{eq:xpar_objective}) and prove that it minimizes the same worst-case loss.
      \textbf{(Middle Right)} When only a single proxy $W$ is used in place of $A$, the result is a weaker guarantee, in the form of a smaller robustness set $C_W(\lambda_1)$ (cf.,~\eqref{eq:single_proxy_ar_guarantee}) for the same value of $\lambda_1$. The shape of this set depends on the noise in the proxy along different dimensions. 
      \textbf{(Far Right)}
      As a result, there does not generally exist a $\lambda_2$ such that $C_W(\lambda_2) = C_A(\lambda_1)$.  If we choose some $\lambda_2 > \lambda_1$ such that $C_A(\lambda_1) \subset C_W(\lambda_2)$, we enforce a stronger constraint than intended, resulting in an unwanted trade-off between performance and robustness.
    }\label{fig:scatter-intervention}
\end{figure*}
First, we establish the robustness set of Anchor Regression when a single proxy is used in place of $A$.  We refer to this as Proxy Anchor Regression, to distinguish it from the case when $A$ is observed, but the only difference from Definition~\ref{def:ar_objective} is that $W$ is used in place of $A$.
\begin{thmdef}[Proxy Anchor Regression]\label{def:par_objective}
Let $\ell_{LS}, \ell_{PLS}$ be defined as in Equations~\eqref{eq:def_ls} and~\eqref{eq:def_pls_linear}.
We define
\begin{align}
  \ell_{PAR}(W; \gamma, \lambda) &\coloneqq \ell_{LS}(\gamma) + \lambda \ell_{PLS}(W; \gamma), \label{eq:par_objective}
\end{align}
where $\lambda \geq -1$ is a hyperparameter and we suppress the dependence on $X, Y$ in the notation. 
\end{thmdef}

\begin{thmthm}\label{thmthm:single_proxy_ar_guarantee} 
Under Assumptions~\ref{asmp:linear_scm} and~\ref{asmp:ar_scm_one_proxy}, for all $\gamma \in \R^{d_X}$ and for all $\lambda \geq -1$
\begin{align*}
  \ell_{PAR}(W; \gamma, \lambda) &= \sup_{\nu \in C_W(\lambda)} \E_{do(A \coloneqq \nu)}[{(Y - \gamma^\top X)}^2],
\end{align*}
where the robustness set is given by 
\begin{equation}
C_{W}(\lambda) \coloneqq \{ \nu: \E[\nu \nu^\top] \preceq \E[AA^\top] + \lambda \Omega_W \}\label{eq:single_proxy_ar_guarantee}
\end{equation}
and where $\Omega_W$ is defined as
\begin{align}
\Omega_W \coloneqq \E[AW^\top] {\left(\E[WW^\top]\right)}^{-1} \E[WA^\top].\label{eq:def_omega_w}
\end{align}
\end{thmthm}
Intuitively, $\Omega_W$ defines a signal-to-variance relationship in $W$, and this determines the robustness guarantee.  In the case where both $A, W \in \R$ are one-dimensional, and $A$ has unit variance, the robustness sets simplify to 
\begin{align*}
  C_{OLS} &= \{ \nu: \E[\nu^2] \leq 1 \}\\
  C_W(\lambda) &= \{\nu: \E [ \nu^2 ] \leq 1 + \lambda \cdot \rho_W \}\\
  C_A(\lambda) &= \{\nu: \E[ \nu^2 ] \leq 1 + \lambda \},
\end{align*}
where $\rho_W \coloneqq \beta_W^2 / (\beta_W^2 + \E \epsilon_W^2) < 1$ is the signal-to-variance ratio of $W$, also referred to as the reliability ratio in the measurement error literature \citep{Fuller1987}.  Thus, in the one-dimensional case, the robustness set using $W$ is strictly smaller than the one obtained by using $A$ when $\lambda > 0$, except in the case where $\epsilon_W = 0$ a.s. This result generalizes to higher dimensions.
\begin{thmprop}\label{thmprop:subset_guarantee}
Assume Assumptions~\ref{asmp:linear_scm} and~\ref{asmp:ar_scm_one_proxy} and that $\E[\epsilon_W \epsilon_W^\top] \in \R^{d_W \times d_W}$ is positive definite.  Then for $\lambda > 0$
\begin{equation*}
  C_{OLS} \subseteq C_W(\lambda) \subset C_A(\lambda),
\end{equation*}%
and the set $C_W(\lambda)$ increases monotonically when $\E[\epsilon_W\epsilon_W^\top]$ decreases w.r.t.\ the partial matrix ordering.  If $d_W = d_A$, $\beta_W$ is full rank, and $\epsilon_W = 0$ a.s., then $C_W(\lambda) = C_A(\lambda)$.
\end{thmprop}
If $\Omega_W$ were known, we could choose a larger $\lambda^*$ such that $C_A(\lambda) \subseteq C_W(\lambda^*)$.  
In contrast to the one-dimensional case, where we could choose $\lambda^* = \lambda / \rho_W$ to obtain an equality $C_A(\lambda) = C_W(\lambda^*)$, we cannot generally achieve equality in higher dimensions (see Figure~\ref{fig:scatter-intervention}).

However, $\Omega_W$ is not generally identifiable from the observed distribution over $(X, Y, W)$ alone.  Moreover, SCMs compatible with the observed distribution react differently under interventions on $A$ and yield different coefficients that are optimal w.r.t.\ interventions in $C_A(\lambda)$.
Consequently, in this setting, it is not possible to recover the guarantees of Anchor Regression without further assumptions (e.g., on $\Omega_W$). See Supplement~\ref{sub:identification_example} for an example.

Note that these results apply regardless of whether or not $\beta_W$ is full rank. However, if $\beta_W$ is not full rank, then there will be directions of variation in $A$ that are not reflected in $W$, and we will not be able to achieve additional robustness (beyond that of OLS) against interventions along these directions.  

\subsection{Robustness with two proxies}%
\label{sec:cross_estimator}

We now show that if we have two (sufficiently different) proxies for $A$, then it is possible to recover the original robustness set using a different regularization term.  We denote these proxies by $W, Z$, as shown in Figure~\ref{fig:causal_graphs}.  In this setting, the structural causal model over $(X, Y, H, A)$ can still be written in the form of Equation~(\ref{eq:ar_scm_full}), where we make the following additional assumptions.
\begin{thmasmp}[Proxies with additive noise]\label{asmp:ar_scm_two_proxy}
In the context of Assumption~\ref{asmp:linear_scm}, $Z, W$ are generated as follows
\begin{align*}
  W &\coloneqq \beta_W^\top A + \epsilon_W &\text{and}&&  Z &\coloneqq \beta_Z^\top A + \epsilon_Z,
\end{align*}
where $\epsilon_W, \epsilon_Z$ are mean-zero with bounded covariance, and $\epsilon_W, \epsilon_Z, \epsilon, A$ are jointly independent.
\end{thmasmp}%
\begin{thmasmp}\label{asmp:zw_invert}
The dimensions of $A, W, Z$ are equal, $d_A = d_W = d_Z$, and $\beta_W, \beta_Z$ are full-rank.
\end{thmasmp}%
Note that Assumption~\ref{asmp:zw_invert} also implies that the second moment matrix $\E[ZW^\top]$ is invertible.  

To build intuition, note that this assumption is trivially satisfied in the setting where $W = A + \epsilon_W$ and $Z = A + \epsilon_Z$, i.e., where $W$ and $Z$ are two noisy observations of $A$.  More generally,  Assumption~\ref{asmp:zw_invert} rules out directions of variation in $A$ that are undetectable in $W$ or $Z$.  


In this setting we introduce the following loss, and prove that it is equal to the worst-case loss obtained when $A$ is observed (c.f.,~\eqref{eq:arguarantee})
\begin{thmdef}[Cross-Proxy Anchor Regression]
\begin{align}
 \ell_{\times PAR}(W, Z; \gamma, \lambda) &\coloneqq \ell_{LS}(X, Y; \gamma) + \lambda \ell_{\times}(W, Z; \gamma), \nonumber
\end{align}
where we refer to 
\begin{equation}
\ell_{\times}(W, Z; \gamma) \coloneqq {\E[{R(\gamma)} W^\top]} {\E[ZW^\top]}^{-1} {\E[Z {R(\gamma)}^\top]},\label{eq:cross_regularization}
\end{equation}
as the cross-proxy regularization term.
\end{thmdef}
\begin{thmthm}\label{thmthm:cross_equals_iv}
Under Assumptions~\ref{asmp:linear_scm},~\ref{asmp:ar_scm_two_proxy} and~\ref{asmp:zw_invert}, for any $\gamma \in \R^{d_X}$ and any $\lambda \geq -1$
\begin{equation}
\ell_{\times PAR}(W, Z; \gamma, \lambda) = \sup_{\nu \in C_{A}(\lambda)} \E_{do(A \coloneqq \nu)}[{(Y - \gamma^\top X)}^2], \label{eq:xpar_objective}
\end{equation}
where $C_{A}(\lambda) = \{ \nu: \E[\nu \nu^\top] \preceq (1 + \lambda) \E[AA^\top]\}$.
\end{thmthm}
$\ell_{\times PAR}$ is convex in $\gamma$ and has a closed form solution for its minimizer
based only on the population moments of $X, Y, W$ and $Z$ (see Proposition~\ref{thmprop:xpar_closed_form} in the supplement).

To build intuition for why Assumption~\ref{asmp:zw_invert} is required for this result, consider an example where $W, Z$ are both scalars ($d_W = d_Z = 1$) and $A$ has two independent dimensions $(A_1, A_2)$.  In this example, if both proxies measure the same dimension $A_1$, then variation in $A_2$ is not detectable in either proxy, and we cannot optimize for robustness to interventions on $A_2$.  On the other hand, if $W$ only measures $A_1$ (e.g., $W = A_1 + \epsilon_W$), and $Z$ only measures $A_2$ (e.g., $Z = A_2 + \epsilon_Z$), then we cannot use $Z$ to identify the signal-to-variance ratio of $W$, and vice-versa.  In this case, $(W, Z)$ is effectively a single two-dimensional proxy in the framework of Section~\ref{sub:a_single_proxy_is_insufficient_for_robustness}, where we showed that recovering the guarantees of Anchor Regression is not generally possible.  
Intuitively, we need all directions of variation in $A$ to have some influence on both proxies (i.e., $\beta_W, \beta_Z$ full rank), and hence require that $W, Z$ have sufficiently large dimension.

\section{Targeted anchor regression: Incorporating additional shift information}%
\label{sec:targeted_shift}

We now generalize Anchor Regression to an estimator that is targeted to be robust against particular shifts, and demonstrate that we can similarly handle this setting when only proxies of $A$ are observed.  In Section~\ref{subsec:dist-robust-AR} we showed that Anchor Regression minimizes the worst-case loss over the set $C_A(\lambda)$ of all interventions $do(A\coloneqq\nu)$ where $\E[\nu\nu^\top]\preceq (1+\lambda)\E[AA^\top]$.
For deterministic $\nu$, $C_A(\lambda)$ is an ellipsoid centered at $0$, and its width in each direction is proportional to the variation of $A$ in that direction.
However, we may desire a different robustness set:  For instance, if we anticipate a particular shift $\mu_\nu$ in the mean of $A$, or if we want to add extra protection against particular directions of variation in $A$.  This can be formalized as a robustness set defined by an ellipsoid that may not be centered at $0$, nor be proportional to $\E[AA^\top]$. The estimator developed in this section can incorporate such prior beliefs.

\begin{figure}[t]
  \centering
  \input{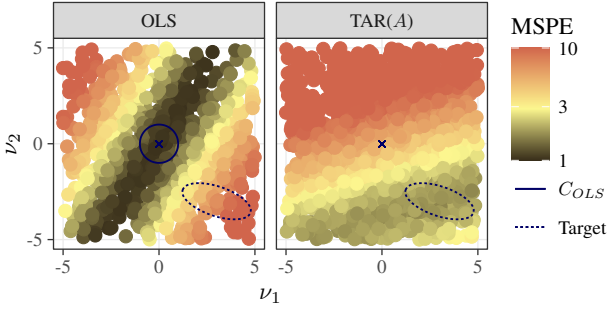}
  \vspace{-20pt}
  \caption{
    Targeted Anchor Regression allows for minimizing the worst-case loss in regions (dashed ellipse) that may differ in location, size, and shape from the regions in Figure~\ref{fig:scatter-intervention} (OLS copied for reference). Every point $\nu$ represents a test distribution $do(A:=\nu)$, the color indicating the mean squared prediction error in this distribution.  Cross marks the origin. The TAR estimator achieves its minimal test loss at the center of the targeted region.
  }\label{fig:scatter-targeted-shift}
\end{figure}

More formally, instead of considering robustness against interventions $do(A \coloneqq \nu)$ over the set $\nu \in C_{A}(\lambda)$, we now assume that we have additional information on the nature of $\nu$, which is specified in the form of a vector $\mu_{\nu}$ and a symmetric PSD matrix $\Sigma_{\nu}$.  We introduce a new method, Targeted Anchor Regression, minimizing what we refer to as the \textit{targeted loss}. We prove in Propositions~\ref{thmprop:targeting_known_shift} and~\ref{thmprop:generalized_ar} that minimizing this objective can be interpreted in two ways: First, as minimizing an expected loss over interventions $\nu$ with a known mean and covariance, or minimizing a worst-case loss over deterministic interventions $\nu$ contained in an ellipsoid robustness set (as discussed above).  This is visualized in Figure~\ref{fig:scatter-targeted-shift}.

\subsection{Targeting when $A$ is observed}%
\label{sub:targeted_shift_obs}

We first consider the case when $A$ is observed during training, and the mean and covariance of $\nu$ are known, given by $\mu_{\nu}, \Sigma_{\nu}$.  Importantly, for a given $\gamma$ we have $\E[R(\gamma) \mid A = a] = b_{\gamma}^\top a$, where, writing $\Sigma_A \coloneqq \E[AA^\top]$,
\begin{equation}
  b_{\gamma}^\top \coloneqq \E[R(\gamma)A^\top] \Sigma_A^{-1}. \label{eq:def_b_gamma}
\end{equation}
\begin{thmdef}[Targeted Anchor Regression]\label{thmdef:tar}
Let $\mu_{\nu} \in \R^{d_A}$, and $\Sigma_{\nu} \in \R^{d_A \times d_A}$, where $\Sigma_{\nu}$ is a symmetric PSD matrix.
\begin{align}
 &\ell_{TAR}(A; \mu_{\nu}, \Sigma_{\nu}, \gamma, \alpha) \nonumber \\
 &\coloneqq \ell_{LS}(\gamma) + b_{\gamma}^\top \left(\Sigma_{\nu} - \Sigma_{A}\right) b_{\gamma} + {(b_{\gamma}^\top \mu_{\nu} - \alpha)}^2, \label{eq:tar_a}
\end{align}
where $b_{\gamma}$ is defined in~\eqref{eq:def_b_gamma}, and $\Sigma_A$ is the covariance of $A$.
\end{thmdef}

\begin{thmprop}\label{thmprop:targeting_known_shift}
Under Assumption~\ref{asmp:linear_scm}, and the assumption that $\nu \indep \epsilon$, 
we have, for all $\gamma \in \R^{d_X}, \alpha \in \R$, 
\begin{equation*}
  \ell_{TAR}(A; \mu_{\nu}, \Sigma_\nu; \gamma, \alpha) = \E_{do(A\coloneqq\nu)}[{(Y - \gamma^\top X - \alpha)}^2],
\end{equation*}
where $\mu_{\nu} = \E[\nu]$ and $\Sigma_{\nu}$ is the covariance matrix of $\nu$.
\end{thmprop}
Importantly, the objective in Equation~\eqref{eq:tar_a} is convex in $(\gamma, \alpha)$,
and has a closed-form solution (see Proposition~\ref{thmprop:convexity_known_shifts} in the supplement).  If $\nu$ is a known constant, then this corresponds to performing OLS using both $X$ and $A$ as predictors during training, and using the known value of $\nu$ for $A$ for prediction (see Supplement~\ref{ssub:proofs_section_target}). However, if for example $\nu$ exhibits more variance than $A$ along certain directions, and less variance along others, then the targeted regression parameter differs from standard solutions. 
Optimizing the objective in Equation~\eqref{eq:tar_a} can also be interpreted as optimizing a worst-case loss over interventions $do(A \coloneqq \nu)$ in a certain set.
\begin{thmprop}\label{thmprop:generalized_ar}
Under Assumption~\ref{asmp:linear_scm}, we have, for all $\mu_{\nu} \in \R^{d_A}$ and $\Sigma_{\nu} \in \R^{d_A \times d_A}$ being a symmetric positive definite matrix, that
\begin{align*}
  &\argmin_{\gamma, \alpha} \ell_{TAR}(A; \mu_{\nu}, \Sigma_{\nu}, \gamma, \alpha) \\
  &\quad \quad = \argmin_{\gamma, \alpha} \sup_{\nu \in T(\mu_{\mu}, \Sigma_{\nu})} \E_{do(A \coloneqq \nu)}[{(Y - \gamma^\top X - \alpha)}^2],
\end{align*}
where the supremum is taken over (deterministic or random) shifts $\nu$ of the form $\nu = \mu_v + \delta$, where $\delta$ satisfies the constraint that $\E[\delta \delta^\top] \preceq \Sigma_\nu$.  If $\delta$ is random, we require that it is independent of all other random variables. In other words, we can write that $\nu$ lies in the set
\begin{equation*}
  T(\mu_{\nu}, \Sigma_{\nu}) \coloneqq \{ \nu: \E[{(\nu - \mu_{\nu})}{(\nu - \mu_{\nu})}^\top] \preceq \Sigma_\nu \}.
\end{equation*}
\end{thmprop}
Note that the expectation in the constraint $T$ is with respect to the random variable $\nu$. This covers the case in which $\nu$ (and hence $\delta$) is deterministic, in which case it is equal to a fixed value with probability one. 

Proposition~\ref{thmprop:generalized_ar} shows that Targeted Anchor Regression generalizes Anchor Regression to a broader class of robustness sets, that need not depend explicitly on $\E[AA^\top]$. In particular, Anchor Regression can be viewed as a special case, where $\Sigma_\nu = (1 + \lambda) \Sigma_A$ and $\E[\nu] = 0$, in which case the objectives are equal for $\alpha = 0$. 
In the following, we adopt the interpretation of $\mu_{\nu}, \Sigma_{\nu}$ as specifying a mean and covariance of $\nu$ (Proposition ~\ref{thmprop:targeting_known_shift}).

\subsection{Targeting with proxies}%
\label{sub:targeted_shift_two_proxies}

In the single-proxy setting, we define Proxy Targeted Anchor Regression as using $W$ in place of $A$ in Equation~\eqref{eq:tar_a}.  We assume a known mean and covariance of $W$ under $\P_{do(A \coloneqq \nu)}$, used in place of $\mu_{\nu}, \Sigma_{\nu}$.  By similar arguments to those in Section~\ref{sub:a_single_proxy_is_insufficient_for_robustness}, this approach does not generally yield the optimal predictor, in a way that depends on the (unidentified) signal-to-variance relationship in $W$.  Given the similarity, we defer details to Supplement~\ref{sub:targeted_shift_one_proxy}.

When two proxies $W, Z$ are available, we can recover the statement from  Proposition~\ref{thmprop:targeting_known_shift} using a modified estimator, by similar arguments to those in Section~\ref{sec:cross_estimator}.  The core observation is that we can construct a linear term
\begin{align}
  a_{\gamma}^{\top} &\coloneqq \E[R(\gamma) Z^\top] {(\E[WZ^\top])}^{-1} \label{eq:def_a_gamma},
\end{align}
which, if $\beta_Z = \beta_W = \operatorname{Id}$ can be seen as a linear IV estimate of $b_{\gamma}^\top$ in Equation~\eqref{eq:def_b_gamma}, an estimator used in the measurement error literature given repeated noisy measurements of a single variable \citep{Fuller1987}. In our case, Equation~\eqref{eq:def_a_gamma} identifies $b_{\gamma}^\top$ only up to the linear transformation $\beta_W$, but this is sufficient to identify the targeted loss.

\begin{thmdef}[Cross-Proxy Targeted Anchor Regression]
Let $\tilde{\mu} \in \R^{d_W}$, and $\tilde{\Sigma}_{W} \in \R^{d_W \times d_W}$, where $\tilde{\Sigma}_{W}$ is a symmetric positive semi-definite matrix.  We define
\begin{align*}
  &\ell_{\times TAR}(W, Z; \tilde{\mu}, \tilde{\Sigma}_{W}, \gamma, \alpha) \\
  &\coloneqq \ell_{LS}(\gamma) + a_{\gamma}^\top \left(\tilde{\Sigma}_{W} - \Sigma_W \right) a_{\gamma} + {\left(a_{\gamma}^\top \tilde{\mu} - \alpha \right)}^2,
\end{align*}
where $a_{\gamma}$ is defined in~\eqref{eq:def_a_gamma}.
\end{thmdef}

In Theorem~\ref{thmthm:two_proxies_target} (Supplement~\ref{sub:targeted_shift_one_proxy}) we prove, analogous to Theorem~\ref{thmthm:cross_equals_iv}, that this population objective is equal to that of Targeted Anchor Regression~\eqref{eq:tar_a}.

\section{Synthetic experiments}%
\label{sec:experiments}
In Section~\ref{sub:exp-robustness-guarantee}, we show that Cross-Proxy Anchor Regression (xPAR) outperforms Proxy Anchor Regression (PAR) in settings with noisy proxies.  As the noise increases, xPAR continues to match Anchor Regression (AR) test performance under intervention, while PAR approaches OLS.  In Section~\ref{sub:exp-misspecified-svr}, we demonstrate the risks of attempting to correct for this noise by assuming a certain signal-to-variance ratio.  In Section~\ref{sub:exp-anticausal-causal} we demonstrate another benefit of xPAR over PAR, giving an example where it places more weight on causal predictors relative to PAR.  Finally, in Section~\ref{sub:exp-target-shift}, we highlight the trade-off between using Targeted Anchor Regression (TAR) vs. OLS and AR, showing that TAR improves performance under the targeted shift, at the cost of incurring additional error on the training distribution.  Code for experiments is available at \url{https://github.com/clinicalml/proxy-anchor-regression}.

\subsection{Mean squared prediction error under intervention}\label{sub:exp-robustness-guarantee}
We demonstrate on synthetic data that xPAR recovers similar test performance to AR, while the performance of PAR degrades as the signal-to-variance ratio (SVR) of the proxies decreases.  We simulate training data (at different levels of signal-to-variance) from an SCM with the structure given in Figure~\ref{fig:causal_graphs}, fix $\lambda \coloneqq 5$ and fit PAR and xPAR.  We then choose a fixed intervention $\nu$, and simulate test data under the intervened distribution, evaluating our learned predictors.

In Figure~\ref{fig:loss-robustness}, we see that the test errors for xPAR and AR coincide (see Theorem~\ref{thmthm:cross_equals_iv}) while PAR interpolates between OLS and AR, depending on the signal-to-variance ratio (see Proposition~\ref{thmprop:subset_guarantee}). Section~\ref{sec:synth_details} gives additional implementation details on this and remaining experiments.
\begin{samepage}
\begin{figure}[t]
    \centering
    \input{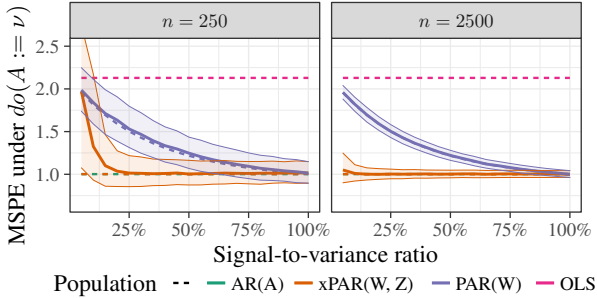}
    \vspace{-5pt}
    \caption{Mean squared prediction error (MSPE) under interventions $do(A \coloneqq \nu)$ for estimators PAR and xPAR. We display population losses for the population parameters as dashed lines, and median empirical MSPE when fit from data as solid lines, with shaded regions covering the 25\% to 75\% quantiles.}
    \label{fig:loss-robustness}
\end{figure}%

\subsection{Misspecified signal-to-variance ratio}\label{sub:exp-misspecified-svr}%
\end{samepage}
In Section~\ref{sub:a_single_proxy_is_insufficient_for_robustness}, we noted that if the (unidentified) signal-to-variance ratio (SVR) were known, we could correct for it when using PAR with a single proxy.  Here we demonstrate the implications of incorrectly specifying this correction.   We simulate data from the same SCM as in Section~\ref{sub:exp-robustness-guarantee}, with varying (true) signal-to-variance ratio.  
\begin{figure}[t]
    \centering
    \input{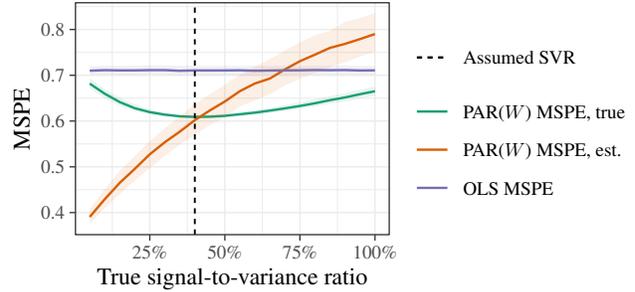}
    \vspace{-15pt}
    \caption{Estimates of worst-case mean squared prediction error (MSPE) over a robustness set $C$. PAR is applied assuming that the signal-to-variance ratio is 0.4, which gives an estimate of the worst-case MSPE over $C$ (orange). Green line shows actual worst-case MSPE over $C$ at different underlying signal-to-variance ratios.}
    \label{fig:misspecified}
\end{figure}

In Figure~\ref{fig:misspecified}, for the predictor chosen by PAR, we plot the estimated worst-case MSPE (in orange), using a correction factor assuming that the signal-to-variance ratio is 0.4, against the true worst-case MPSE (in green).  We observe that if the true signal-to-variance ratio is smaller than our assumption of 0.4, then our estimate is too conservative, and vice versa if the true signal-to-variance ratio is larger.

\subsection{Causal and anti-causal predictors}\label{sub:exp-anticausal-causal}
We demonstrate the ability of xPAR to select causal predictors, in a synthetic setting where predictors $X$ may contain both causal and anti-causal predictors.  We simulate data from an SCM (Figure~\ref{fig:scm} [top]), where one anchor, $A_1$, is a parent of the causal predictors, while the other, $A_2$, is a parent of the anti-causal predictors.  We consider two identically distributed noisy proxies $W, Z$ of $A\coloneqq (A_1, A_2)$.  The challenge is that $A_2$ is measured with significantly more noise than $A_1$, across both proxies. 

As seen in Figure~\ref{fig:scm} [bottom] PAR places more weight on anti-causal features.  In effect, the noise in the measurement of $A_2$ causes $X_{\text{anti-causal}}$ to appear less sensitive to shifts in $A_2$. 
This is an ideal scenario for xPAR, as it is designed to deal with additional noise by leveraging both proxies.  Consequently, when two proxies $W, Z$ are available, xPAR
places more weight on the causal predictors, relative to PAR.

\subsection{Targeted shift}\label{sub:exp-target-shift}
We demonstrate the trade-off made by Targeted Anchor Regression (TAR) versus Anchor Regression (AR), considering the case when $A$ is observed for simplicity.  We simulate training data and fit estimators $\gamma_{\text{OLS}}$, $\gamma_{\text{AR}}$ and $\gamma_{\text{TAR}}$, where $\gamma_{\text{TAR}}$ is targeted to a particular mean and covariance of a random intervention $\nu$, and we select $\lambda$ for $\gamma_{\text{AR}}$ such that this intervention is contained within $C_A(\lambda)$.

\begin{figure}[t]
  \centering
  \input{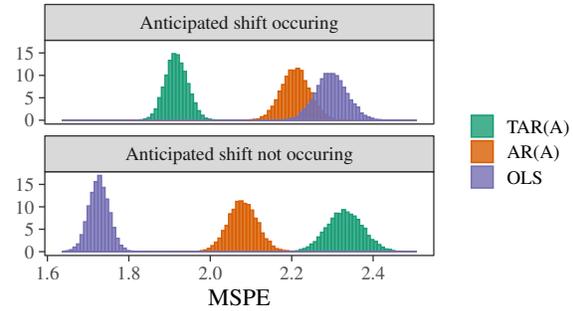}
  \vspace{-10pt}
  \caption{Empirical mean squared prediction error of TAR, OLS and AR\@ under the shifted distribution and the training distribution.}%
  \label{fig:targeted-shift}
\end{figure}

We then simulate test data from two distributions:  $\P_{do(A \coloneqq \nu)}$ (i.e., the shift occurs), and $\P$ (where it does not), and evaluate the mean squared prediction error (MSPE). The results are shown in Figure~\ref{fig:targeted-shift}, and demonstrated that TAR performs better than AR and OLS in the first scenario, but this comes at the cost of worse performance on the training distribution.

\section{Real-data experiment: Pollution}%
\label{sec:experiment_pollution}

We test our approach on a real-world heterogeneous dataset of hourly pollution readings in five cities in China, taken over several years \citep{Liang2016}, with most data available from 2013-15.  Our prediction target is PM2.5 concentration, a measure of pollution, and covariates are primarily weather-related, including dew point, temperature, humidity, pressure, wind direction / speed, and precipitation.  

\begin{figure}[t]
\captionsetup[subfigure]{labelformat=empty}
\centering
\begin{subfigure}[t]{\linewidth}
    \centering
    \begin{tikzpicture}[
      obs/.style={circle, draw=gray!90, fill=gray!30, very thick, minimum size=7mm},
      uobs/.style={circle, draw=gray!90, fill=gray!10, dotted, minimum size=7mm},
      none/.style={rectangle, draw=gray!0, fill=gray!0, dotted, minimum size=7mm},
      bend angle=25]
      \node[none] (U1) at (0, 1) {$A_1$};
      \node[none] (U2) at (0, 0) {$A_2$} ;
      \node[none] (X1) [right=of U1] {$X_{\text{causal}}$} ;
      \node (X1circ) at (X1) {};
      \node[none] (X2) [right=of U2] {$X_{\text{anti-causal}}$} ;
      \node (X2circ) at (X2) {};
      \node[none] (W) [left=of U1] {$W$};
      \node[none] (Z) [left=of U2] {$Z$} ;
      \node[none] (Y) [right=of X1, yshift = -5mm] {$Y$};
      \draw[-latex, thick] (U2) -- (X2);
      \draw[-latex, thick] (U1) -- (X1);
      \draw[-latex, thick] (U1) -- (W);
      \draw[-latex, dotted] (U2) -- (Z); 
      \draw[-latex, thick] (U1) -- (Z);
      \draw[-latex, dotted] (U2) -- (W); 
      \draw[-latex, thick] (X1) -- (Y);
      \draw[-latex, thick] (Y) -- (X2);
    \end{tikzpicture}
    \caption{} 
    \label{fig:scm-fig-sub1}
\end{subfigure}
\begin{subfigure}[t]{\linewidth}
    \centering
    \input{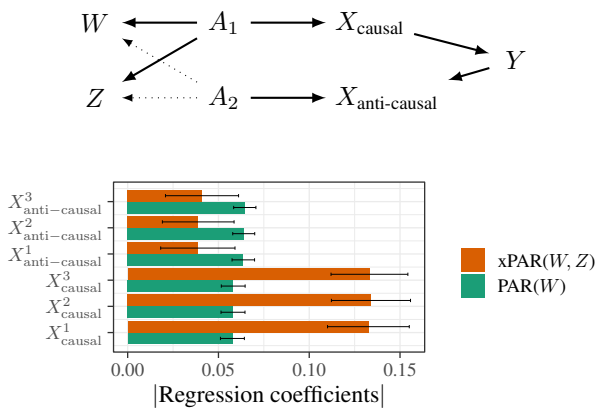}
    \caption{}
    \label{fig:scm-fig-sub2}
\end{subfigure}
\vspace{-35pt}
\caption{\textit{Top}: SCM with $A_1, A_2$ (unobserved), target $Y$ and predictor variables $X_{\text{causal}},  X_{\text{anti-causal}} \in \R^{3}$. Dotted lines indicate higher noise. \textit{Bottom}: Absolute value of regression coefficients. PAR places more weight on anti-causal predictors, while xPAR places more weight on causal predictors.}
\label{fig:scm}
\end{figure}

\textbf{Real-World Proxy (Temperature)}: Pollution tends to be seasonal in this dataset, and so we construct our training and test environments using seasons:  For each of the four seasons, we train only on the other three seasons, and evaluate on the held-out season.  We do this for each city, treating each city and held-out season as a separate evaluation.  This leads to 20 separate scenarios.  

With this variation in mind, we use temperature as a real-world proxy, and treat it as unavailable at test time. We also construct two noisier copies of temperature, which we refer to as $W, Z$, adding independent Gaussian noise while controlling the signal-to-variance ratio (in the training distribution) at $\var(\text{Temp}) / \var(W) = 0.9$.

\textbf{Estimators / Benchmarks}: For Proxy and Cross-Proxy AR (PAR, xPAR, see Section~\ref{sec:worst_case_bounded_shift}), we choose $\lambda \in [0, 40]$ by leave-one-group-out cross-validation on the three training seasons, using the first year (2013) of data.  For instance, if \enquote{winter} is the test season, then we choose the value of $\lambda$ that performs best on average across combinations of the other seasons e.g., training on the fall \& summer data and evaluating on the spring data.

When using temperature as a single proxy in PAR, we observe that in 9 out of 20 scenarios, $\lambda = 40$ is chosen, but in the remaining 11, $\lambda = 0$ is chosen, which is equivalent to OLS\@. For comparability, we use the same values of $\lambda$ for PAR$(W)$ and xPAR$(W, Z)$. For Proxy Targeted AR and Cross-Proxy Targeted AR (PTAR, xPTAR, see Section~\ref{sec:targeted_shift}), we use the mean and variance of the relevant variables (e.g., temperature, $W$, $Z$) in the held-out season to target our predictors.  

Our primary benchmark is OLS (without temperature).  We also compare to (a) OLS that uses temperature during train and test [OLS (TempC)], and (b) OLS that includes the temperature during training, and uses the mean test value for temperature during prediction [OLS + Est. Bias].  We present the results for the 9 scenarios where $\lambda > 0$ in Table~\ref{tab:mse_pollution}, since PAR with $\lambda=0$ is equivalent to OLS (aggregate results in Table~\ref{tab:mse_pollution_appendix} in the supplement).  

\textbf{Results}: For both PAR and PTAR, we see improvement over OLS on average across scenarios, with limited downside (e.g., in the worst scenario for PTAR relative to OLS, the additional MSE incurred is 0.001). 
In Figure~\ref{fig:comparison_coefficients} (Supplement), we observe that PAR and PTAR achieve gains in two different ways: PAR increases the coefficients of humidity and dew point relative to OLS, while PTAR reduces them and incorporates a correction into the intercept.

\begin{table}[t]
  \centering
  \caption{Mean: Average MSE (lower is better) over 9 scenarios where $\lambda > 0$. \# Win: Number of scenarios where the estimator has lower MSE than OLS. Best (Worst): Smallest (Largest) difference to OLS across environments, where lower is better.}
  \label{tab:mse_pollution}
\begin{tabular}{lrrrr}
\toprule
Estimator &  Mean &  \# Win &  Best & Worst\\
\midrule
OLS             & 0.537 &           &  & \\
OLS (TempC)     & 0.536 &    5       & -0.028 & 0.026 \\
OLS + Est. Bias & 0.569 &    4       & -0.072 & 0.150 \\
\midrule
PAR (TempC)     & 0.531 &    6       & -0.041 & 0.006 \\
PAR (W)         & 0.531 &    6       & -0.037 & 0.006 \\
xPAR (W, Z)     & 0.531 &    6       & -0.039 & 0.007 \\
\midrule
PTAR (TempC)    & 0.525 &    8       & -0.061 & 0.001 \\
PTAR (W)        & 0.529 &    8       & -0.038 & 0.001 \\
xPTAR (W, Z)    & 0.526 &    7       & -0.059 & 0.001 \\
\bottomrule
\end{tabular}  
\end{table}

\section{Discussion and related work}%
\label{sec:discussion_and_related_work}

Learning a predictive model that performs well under arbitrarily strong causal interventions is an ambitious goal.  In this work, we have argued that even if causal invariance is achievable, it may not be desirable: A model whose performance is invariant to arbitrarily strong interventions may have poor performance when the test distribution does not differ too much from the training distribution.

There is a large body of work that seeks to learn causal models as a route to achieving invariance \citep{Rojas-Carulla2015, Magliacane2017}, or that uses knowledge of the causal graph to select predictors with invariant performance under a set of known interventions \citep{Subbaswamy2019}. 
Similarly, invariant risk minimization (IRM) seeks a predictor $\Phi$ such that $\E(Y \mid \Phi(X))$ is invariant across a set of discrete environments \citep{Arjovsky2019-kv, Xie2020-fn, Krueger2020-np, Bellot2020-dd}. Recent work has pointed to the theoretical and practical difficulty of learning such a predictor for IRM \citep{Rosenfeld2020, Kamath2021-ww, Guo2021-bn}, in part due to the fact that recovering a truly invariant model, even in linear settings, requires a large number of environments. Generalization in non-linear settings requires sufficient overlap between environments and strong restrictions on the model class \citep[e.g.,][]{Christiansen2020}. In contrast to all of the above, we trade off between in-distribution performance and invariance explicitly, instead of seeking invariance as a primary goal. Moreover, since we allow for $A$ to influence $Y$ directly and through hidden variables, invariance may not even be achievable, but we can still formulate a worst-case loss for bounded interventions.

We argue for incorporating prior knowledge about potential shifts by (1) identifying proxies for relevant factors of variation (i.e., anchor variables), and (2) specifying plausible sets of interventions on these factors of variation.  We build upon the causal framework of Anchor Regression \citep{Rothenhausler2018}, extending it in two important ways. 

To start, we relax the assumption that the anchor variables are directly observed.  Instead, we only assume access to proxies, and prove that identification of the worst-case loss is feasible with two proxies.  The challenge of identifying the worst-case loss is related to the problem of identifying causal effects with noisy proxies of unmeasured confounders \citep{TchetgenTchetgen2020, Miao2018bridge, Shi2018, Kuroki2014}, and the challenge of learning under classical measurement error \citep{Fuller1987, Hyslop2001, Bound2001}.  Our observation that a single proxy will underestimate the worst-case loss is related to the well-known problem of regression dilution bias \citep{Frost2000}, where performing linear regression under measurement error leads to bias in parameter estimation.  In contrast, we are not concerned with causal / structural parameter estimation, which is generally not possible in the models we consider, but rather estimating a worst-case loss under a class of interventions. \citet{Srivastava2020} also consider distributional shift in unmeasured variables for which proxies are available, and apply techniques for handling worst-case sub-populations from DRO \citep{Duchi2020}.  In contrast, we consider causal interventions on $A$ that could lie outside the support of the training data, which cannot be represented as a sub-population.  Moreover, they consider the single-proxy case, and give a generalization bound that incorporates the impact of noise, while under our assumptions we are able to recover guarantees as if $A$ were observed, using two proxies.

We then introduce Targeted Anchor Regression, a method for incorporating additional prior knowledge on the strength and direction of shifts in anchor variables. This method can be interpreted as allowing for specification of a broader class of robustness sets, beyond those considered in \citet{Rothenhausler2018}, or as specifying the mean and covariance of the anchors at test time.  We prove analogous results with proxies in this setting, and evaluate this strategy empirically in Section~\ref{sec:experiment_pollution}, targeting our loss to a particular mean and variance over temperature in the held-out season.  

Our work contributes to a growing body of literature that seeks to generalize Anchor Regression to new settings, whether allowing for unobserved anchors and a broader class of robustness sets (as in our work), or generalizing to discrete and censored outcomes, as in \citet{Kook2021-lk}.

\section*{Acknowledgements}
We thank Hussein Mozannar, Chandler Squires, Hunter Lang, Zeshan Hussain, and other members of the ClinicalML lab for feedback and insightful discussions. This work was supported in part by Office of Naval Research Award No.\@ N00014-17-1-2791. NT and JP are supported by a research grant (18968) from VILLUM FONDEN, and JP, in addition, is supported by Carlsberg Foundation.

\clearpage
\appendix

\section*{Supplementary Materials}
The supplementary materials are organized as follows
\begin{itemize}
\item (Section~\ref{sec:supp_intuition}): First, we give a simple 1D example to build intuition for the theoretical results.
\item (Section~\ref{sub:identification_example}): In the context of Section~\ref{sub:a_single_proxy_is_insufficient_for_robustness}, we give a concrete example to demonstrate the non-identifiability of $\Omega_W$, defined in~\eqref{eq:def_omega_w}.  We focus on the simple case when $W$ is one dimensional, and the matrix $\Omega_W$ reduces to a single number $\rho_W \coloneqq \beta_W^2 / (\beta_W^2 + \sigma_W^2)$, indicating the signal-to-variance ratio of $W$.  We give an example of an observed distribution for which $\rho_W$ is not identified, and moreover, the optimal predictor with respect to the robustness set $C_A(\lambda)$ is not identified (see Figure~\ref{fig:id_example}).
\item (Section~\ref{sec:proofs}): Proofs for results stated in the main paper.
\item (Section~\ref{sub:targeted_shift_one_proxy}): Additional results (and proofs) for Proxy Targeted Anchor Regression (PTAR) and Cross-Proxy TAR, deferred from the main paper.
\item (Section~\ref{sec:synth_details}): Details for implementation of all experiments
\item (Section~\ref{sec:addl_exp_svr}): Additional synthetic experimental results
\end{itemize}

\section{An example for building intuition}
\label{sec:supp_intuition}

To illustrate the problem, consider the following setup, where we observe $A, X, Y$ at training time, and wish to learn a predictor $\hat{y} = \alpha + \gamma x$ that will generalize to a new environment where $\P_{te}(A) \neq \P_{tr}(A)$.

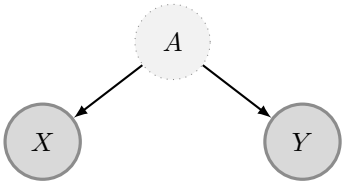
\begin{figure}[ht]
  \centering
    \begin{tikzpicture}[
      obs/.style={circle, draw=gray!90, fill=gray!30, very thick, minimum size=10mm},
      uobs/.style={circle, draw=gray!90, fill=gray!10, dotted, minimum size=10mm},
      bend angle=30]
      \node[uobs] (A) {$A$} ;
      \node[obs] (Y) [below right=of A, yshift=4mm] {$Y$};
      \node[obs] (X) [below left=of A, yshift=4mm] {$X$} ;
      \draw[-latex, thick] (A) -- (X);
      \draw[-latex, thick] (A) -- (Y);
    \end{tikzpicture}
    \caption{Simple example where $X, Y, A \in \R$.}%
    \label{fig:simple_confounder}
\end{figure}
Suppose that our data is generated under $\P_{tr}$  as follows
\begin{align*}
  A &= \epsilon_A, & \epsilon_A \sim \cN(0, 1) \\
  X &= A + \epsilon_X, & \epsilon_X \sim \cN(0, \sigma_X^2) \\
  Y &= A + \epsilon_Y, & \epsilon_Y \sim \cN(0, \sigma_Y^2),
\end{align*}
where $\epsilon_A, \epsilon_X, \epsilon_Y$ are jointly independent. This simple example demonstrates a few concepts:
\begin{itemize}
  \item Assuming $\sigma_X^2 > 0$, the conditional expectation $\E[Y \mid X]$ changes as the distribution of $A$ changes.
  \item We can write the residuals $Y - \hat{Y}$ as a linear function in $A$ and the noise variables. This holds, even if the errors are non-Gaussian.
  \item The test population MSE is a convex function of $\alpha, \gamma$.
\end{itemize}
In particular, we will see that the parameters $\alpha, \gamma$ trade off between the variance of $A$ and $\epsilon_X$: There exists an invariant solution, where $\alpha = 0, \gamma^* = 1$, such that the MSE is completely independent of $A$, but this is only optimal in the setting where $\var(A) \rightarrow \infty$.

\paragraph{Conditional Expectation depends on $A$} Starting with the assumption that $A, X, Y$ are multivariate Gaussian, we can write down the optimal predictor in the target environment, supposing that at test time $\P_{te}(A) \eid \cN(\mu_A, \sigma_A^2)$.
\begin{align*}
  \E_{te}[Y \mid X = x] &= \E_{te}[Y] + \frac{\cov_{te}(X, Y)}{\var_{te}(X)} \cdot (x - \E_{te}[X]) \\
                   &= \mu_A + \underbrace{\frac{\sigma_A^2}{\sigma_A^2 + \sigma_X^2}}_{\gamma} \cdot (x - \mu_A) \\
                   &= \mu_A (1 - \gamma) + \gamma x,
\end{align*}
where if $\epsilon_X = 0$, then $\gamma = 1$ and the optimal solution does not depend on the parameters of $A$, and is given by  
\begin{equation}
  \E_{te}[Y | X = x] = x.
\end{equation} 
However, for any $\sigma_x^2 > 0$, the optimal solution under $\P_{te}(A)$ depends on $\mu_A, \sigma_A^2$.

\paragraph{Rewriting residuals} Regardless of whether the Gaussian assumption holds, for a given predictor $\hat{Y} = \alpha + \gamma x$, we can write the error $Y - \hat{Y}$ as a function that is linear in $A$ and the noise variables
\begin{align*}
  Y - \hat{Y} &= (A + \epsilon_Y) - \gamma (A + \epsilon_X) - \alpha  \\
               &= A (1 - \gamma) + (\epsilon_Y - \gamma \epsilon_X - \alpha).
\end{align*}
\paragraph{Optimizing for a known target distribution}
The mean squared error $\E[{(Y - \hat{Y})}^2]$ can be written as a function of $\alpha, \gamma$, and the mean and variance of $A$ under $\P_{te}(A)$. Here, all expectations are taken with respect to the test distribution.
\begin{align}
  \E_{te}[{(y - \hat{y})}^2] &= \E_{te}[\E_{te}[{(y - \hat{y})}^2 \mid A]] \nonumber \\
  &= \alpha^2 - 2\alpha \E_{te}[A] (1 - \gamma) \nonumber \\
  &\quad + {(1 - \gamma)}^2 \E_{te}[A^2] + \gamma^2 \sigma_x^2 + \sigma_y^2. \label{eq:example_mse}
\end{align}
By first-order conditions, this expression is minimized by
\begin{align}
  \alpha^* &= \mu_A (1 - \gamma^*) & \gamma^* &= \frac{\sigma_A^2}{\sigma_A^2 + \sigma_X^2}. \label{eq:example_optimum} 
\end{align}
When $\sigma_A^2 \rightarrow \infty$, then $\gamma^* \rightarrow 1$ from Equation~\eqref{eq:example_optimum}.  This is intuitive, because in Equation~\eqref{eq:example_mse}, $\gamma = 1$ renders the MSE functionally independent of the distribution of $A$.  

\paragraph{Optimizing for a worst-case distribution}
Equation~\eqref{eq:example_optimum} shows the optimal solution under a known target distribution, if $\mu_A, \sigma_A^2$ were known in advance. However, a similar intuition applies to the case where $\P_{te}(A)$ is unknown, but we expect it to lie in a particular class.  Consider interventions of the form $do(A \coloneqq \nu)$, where we constrain $\nu$ to lie in the set of random variables $C(\lambda) \coloneqq \{ \nu : \E[\nu^2] \leq \lambda \}$.  In this case, our worst-case loss is given by 
\begin{align*}
  &\sup_{\nu \in C(\lambda)} \E_{\nu}[{(Y - \hat{Y})}^2] \\
  =& \sup_{\nu \in C(\lambda)} (1 - \gamma) \left[ -2 \alpha \E[\nu] + (1 - \gamma) \E[\nu^2] \right] \\
   &+ \alpha^2 + \gamma^2 \sigma_X^2 + \sigma_Y^2,
\end{align*}
where the last line does not depend on $\nu$.  We observe that $\alpha^* = 0$, by analyzing two cases.  First, if $\gamma = 1$, then the first term is eliminated, and the only term that depends on $\alpha$ is $\alpha^2$.  Second, if $\gamma \neq 1$, then ${(1 - \gamma)}^2 > 0$, the first term is partially maximized when $\E[\nu^2] = \lambda$, and if $\alpha \neq 0$, then the expression can be made even larger by choosing a deterministic $\nu = \pm \sqrt{\lambda}$ (instead of e.g., a random $\nu \sim \cN(0, \lambda^2)$), depending on the sign of $\alpha (1 - \gamma)$.  From this (and the presence of the $\alpha^2$ term in the second line) it follows that $\alpha^* = 0$,  in this case as well.  When $\alpha = 0$, the supremum is obtained by any random or deterministic $\nu$ such that $\E[\nu^2] = \lambda$.

With $\alpha^* = 0$ and taking $\E[\nu^2] = \lambda$ in the supremum, this expression simplifies to 
\begin{align*}
  &\sup_{\nu \in C(\lambda)} \E_{\nu}[{(Y - \hat{Y})}^2] \\
  &= {(1 - \gamma)}^2 \lambda + \gamma^2 \sigma_X^2 + \sigma_Y^2.
\end{align*}
Differentiating with respect to $\gamma$, we obtain
\begin{equation*}
  \gamma^* = \frac{\lambda}{\sigma_X^2 + \lambda}.
\end{equation*}
Here, $\lambda$ trades off accuracy and stability; As $\lambda \rightarrow \infty$, we recover the solution where $\gamma^* = 1$, but for situations where $\sigma_X^2$ is large and $\lambda$ is bounded, we are better off choosing $\gamma^* < 1$.

\section{Example: Non-identifiability of $\Omega_W$}%
\label{sub:identification_example}

\begin{figure*}[t]
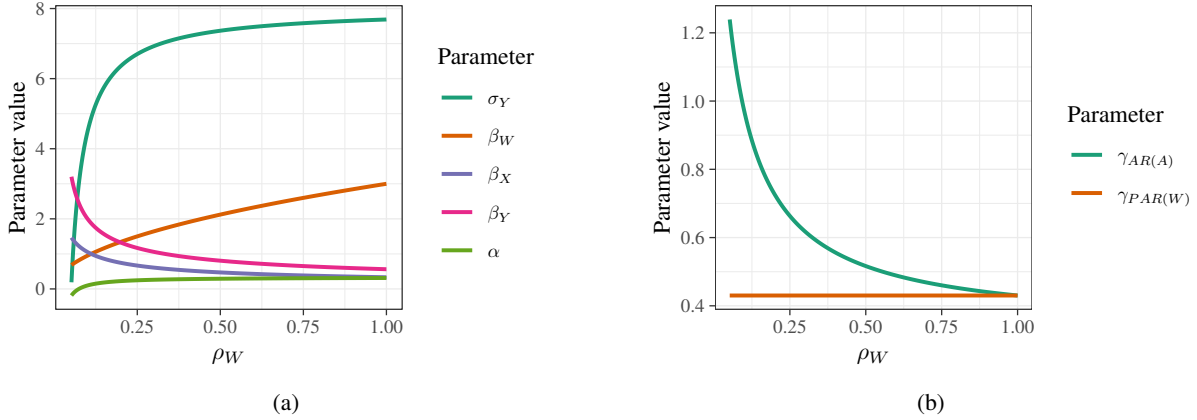

\centering
  \begin{subfigure}[]{0.5\textwidth}
  \centering
  \input{figs/obs-equiv-a-tikz}
  \caption{}\label{fig:id_example_parameters}
  \end{subfigure}%
  \begin{subfigure}[]{0.5\textwidth}
  \centering
  \input{figs/obs-equiv-b-tikz}
  \caption{}\label{fig:id_example_gamma}
  \end{subfigure}
  \caption{(\subref{fig:id_example_parameters}) SCM parameters that all give rise to the same observational distribution, and observe that (\subref{fig:id_example_gamma}) the parameter $\gamma_{AR(A)}$ (as if $A$ were observed) can diverge substantially from the solution $\gamma_{PAR(W)}$, when a single proxy is available. $\lambda = 5$ for this example.
  }
  \label{fig:id_example}
\end{figure*}

\paragraph{Overview} In the context of Section~\ref{sub:a_single_proxy_is_insufficient_for_robustness}, we give a concrete example to demonstrate the non-identifiability of $\Omega_W$, defined in~\eqref{eq:def_omega_w}.  We focus on the simple case when $W$ is one dimensional, and the matrix $\Omega_W$ reduces to a single number $\rho_W \coloneqq \beta_W^2 / (\beta_W^2 + \sigma_W^2)$, indicating the signal-to-variance ratio of $W$.  We give an example of an observed distribution for which $\rho_W$ is not identified, and moreover, the optimal predictor with respect to the robustness set $C_A(\lambda)$ is not identified (see Figure~\ref{fig:id_example}).

\paragraph{Setup} If $(X, Y, W) \in \R^{3}$ is distributed multivariate normal with zero mean, then their covariance matrix fully determines the observed distribution.  Let that covariance matrix be denoted by $\Sigma_{(X, Y, W)} \in \R^{3 \times 3}$, which gives us six observed moments of the distribution
\begin{align*}
  \Sigma_{(X, Y, W)} &\coloneqq  \begin{pmatrix}
    \E[X^2] & \cdot & \cdot \\
    \E[XY] & \E[Y^2] & \cdot \\
    \E[WX] & \E[WY] & \E[W^2]
  \end{pmatrix},
\end{align*}
where we only show the lower triangular portion, since the matrix is symmetric. Suppose that we knew that this observed distribution was generated by the following SCM, but that we do not know the values for the parameters $(\beta_W, \beta_X, \beta_Y, \alpha, \sigma_W^2, \sigma_X^2, \sigma_Y^2)$
\begin{align*}
  A &\coloneqq  \epsilon_A & \epsilon_A \sim \cN(0, 1) \\
  W &\coloneqq  \beta_W A + \epsilon_W & \epsilon_W \sim \cN(0, \sigma_W^2) \\
  X &\coloneqq  \beta_X A + \epsilon_X & \epsilon_X \sim \cN(0, \sigma_X^2) \\
  Y &\coloneqq  \alpha X + \beta_Y A + \epsilon_Y & \epsilon_Y \sim \cN(0, \sigma_Y^2),
\end{align*}
where $\epsilon_A, \epsilon_W, \epsilon_X, \epsilon_Y$ are jointly independent.  We can attempt to identify the parameters using the following relationships implied by the SCM, and matching these to the moments that we observe
\begin{align*}
  \E[WX] &= \beta_W \beta_X \\
  \E[XY] &= \beta_Y \beta_X + \alpha \E[X^2]\\
  \E[WY] &= \beta_W (\beta_Y + \alpha \beta_X) \\
  \E[W^2] &= \beta_W^2 + \sigma_W^2  \\
  \E[X^2] &= \beta_X^2 + \sigma_{X}^2 \\
  \E[Y^2] &= \alpha^2 \E[X^2] + 2\alpha \beta_Y \beta_X  + \beta_Y^2 + \sigma_Y^2
\end{align*}
However, as we will see, this does not identify the parameters.  In particular, there is a set of parameterizations which all give rise to the same observed distribution, and which imply different values of the signal-to-variance ratio $\rho_W \coloneqq \beta_W^2 / (\beta_W^2 + \sigma_W^2)$.

\paragraph{A class of observationally equivalent SCMs} Let $\theta \coloneqq (\beta_W, \beta_X, \beta_Y, \alpha, \sigma_W^2, \sigma_X^2, \sigma_Y^2) \in \R^{7}$ be the parameters of the SCM, and let $\Sigma = f(\theta)$ be the covariance matrix over $(X, Y, W)$ implied by these parameters.

For any covariance matrix $\Sigma$, there exists a subset $C \subset [0, 1]$ such that for any $\rho_W \in C$, we can write the parameters as a function of $\rho_W$, such that $f(\theta(\rho_W)) = \Sigma$.  The set $C$ is constrained by the observed moments:  In particular, as we show below, $\rho_W \geq \text{corr}(W, X)^2$ due to the constraint that $\sigma_X^2 \geq 0$, and the condition that $\sigma_Y^2 \geq 0$ also imposes a lower bound.
In particular, for the covariance matrix below, we demonstrate numerically that $[0.06, 1] \subset C$.
\begin{align*}
  \Sigma_{(X, Y, W)} &\coloneqq  \begin{pmatrix}
    9 & 3 & 1 \\
    3 & 9 & 2 \\
    1 & 2 & 9 
  \end{pmatrix}.
\end{align*}

We now give a strategy for constructing $\theta(\rho_W)$, given a desired $\rho_W$ (including checking the constraint that this $\rho_W \in C$).  Suppose that $W$ and $X$ are positively correlated, as in this example.  Fixing some $\rho_W \in [0, 1]$, we start by writing $\beta_W, \sigma_W$ as functions of $\rho_W$, where 
\begin{align*}
\beta_W &\coloneqq \sqrt{\E[W^2] \rho_W} \\
\sigma_W^2 &\coloneqq \E[W^2](1 - \rho_W).
\end{align*}
The first constraint, that $\sigma_X^2 \geq 0$, can be captured as follows.  Let $\rho_X \coloneqq \beta_X^2 / \E[X^2]$.  Observe that $\sqrt{\rho_X \rho_W} = \text{corr}(W, X)$.  This implies a lower bound on $\rho_W$, given by $\rho_W \geq \text{corr}(W, X)^2$, since $\rho_X \leq 1$ due to $\sigma_X^2 \geq 0$.  This also implies that $\rho_X$ is determined uniquely by $\rho_W$, and is given by $\rho_X = {\text{corr}(W, X)}^2 / \rho_W$.  From this we can write
\begin{align*}
\beta_X &\coloneqq \sqrt{\E[X^2] \rho_X} \\
\sigma_X^2 &\coloneqq \E[X^2] (1 - \rho_X).
\end{align*}
These choices for $(\beta_W, \sigma_W^2, \beta_X, \sigma_X^2)$ match the observed moments $\E[X^2], \E[W^2], \E[WX]$. Then the rest of the parameters can be found as follows, where $\beta_W, \beta_X$ are fixed as above
\begin{align*}
  \beta_Y &\coloneqq \frac{1}{\beta_W (1 - \rho_X)} \left(\E[WY] - \frac{\E[XY]\E[WX]}{\E[X^2]}\right) \\ 
  \alpha &\coloneqq \frac{\E[XY] - \beta_Y \beta_X}{\E[X^2]} \\
 \sigma_Y^2 &\coloneqq \E[Y^2] - \beta_Y^2 - 2\alpha \beta_Y \beta_X - \alpha^2 \E[X^2]
\end{align*}
where all of these are functions of $\rho_W$, in that $\beta_W, \beta_X$ are functions of $\rho_W$.  It remains to verify that for a given choice of $\rho_W$, we satisfy the constraint that $\sigma_Y^2 \geq 0$.  For simplicity, we check this constraint computationally in the context of Example 1, for a range of values of $\rho_W$, and we give the set of observationally-equivalent parameters in Figure~\ref{fig:id_example_parameters}, where valid values of $\rho_W$ range over $[0.06, 1]$.

Next we show that the Proxy Anchor Regression estimator, $\gamma_{PAR(W)}$, differs from the Anchor Regression estimator, $\gamma_{AR(A)}$, and more so when $\rho_W$ becomes small. 
This is shown in Figure~\ref{fig:id_example_gamma}, for $\lambda = 5$, and we give the relevant computations here.

\paragraph{Solution to PAR($W$)}
If we have a single proxy, then we can write down the optimization problem Equation~\eqref{eq:par_objective} as
\begin{align*}
  &\min_{\gamma} \E[{(Y - \gamma X)}^2] + \lambda {\E[(Y - \gamma X) W]}^2 {\E[W^2]}^{-1}\\
  =& \min_{\gamma} \E[Y^2] - 2 \gamma \E[YX] + \gamma^2 E[X^2] \\
   &+ \lambda {(\E[YW] - \gamma \E[XW])}^2 {\E[W^2]}^{-1},
\end{align*}
from which we obtain the optimal solution
\begin{align*}
\gamma_{PAR(W)} &= \frac{\E[YX]\E[W^2] + \lambda \E[YW]}{\E[X^2]\E[W^2] + \lambda \E[XW]}.
\end{align*}
\paragraph{Solution to AR($A$)}
First, we can write the residual as 
\begin{align*}
  &Y - \hat{Y} = Y - \gamma X \\
  &= \alpha X + \beta_Y A + \epsilon_Y - \gamma \beta_X A - \gamma \epsilon_X \\
               &= \alpha (\beta_X A + \epsilon_X) + \beta_Y A + \epsilon_Y - \gamma \beta_X A - \gamma \epsilon_X \\
               &= A ((\alpha - \gamma) \beta_X + \beta_Y) + (\alpha - \gamma) \epsilon_X + \epsilon_Y,
\end{align*}
such that the expected squared error is given by
\begin{align}
  &\E_{do(A \coloneqq \nu)} {(Y - \hat{Y})}^2 \nonumber \\
  &={((\alpha - \gamma) \beta_X + \beta_Y)}^2 \E[\nu^2] + {(\alpha - \gamma)}^2 \sigma_X^2 + \sigma_Y^2,\label{eq:ex1_worst_case_error}
\end{align}
and when $\nu \in \{ \nu: \E[\nu^2] \leq (1 + \lambda) \}$, taking the supremum involves replacing $\E[\nu^2]$ with $(1 + \lambda)$.  Optimizing Equation~\eqref{eq:ex1_worst_case_error} with respect to $\gamma$, we obtain
\begin{align*}
  &\frac{\partial }{\partial \gamma} \left[{((\alpha - \gamma) \beta_X + \beta_Y)}^2 (1 + \lambda) + {(\alpha - \gamma)}^2 \sigma_X^2 + \sigma_Y^2\right] \\
  &= -2 \beta_X (\alpha \beta_X - \gamma \beta_X + \beta_Y)(1 + \lambda) - 2(\alpha - \gamma) \sigma_X^2 ,
\end{align*}
which implies that
\begin{align*}
  0 &= \beta_X (\alpha \beta_X - \gamma \beta_X + \beta_Y)(1 + \lambda)  + (\alpha - \gamma) \sigma_X^2  \\
    &= (\alpha \beta_X^2 + \beta_X \beta_Y)(1 + \lambda) - \gamma \beta_X^2(1 + \lambda) + \alpha \sigma_X^2 - \gamma \sigma_X^2,
    \end{align*}
so that the optimal choice of $\gamma$ is given by 
\begin{equation*}
  \gamma_{AR(A)}= \frac{(\alpha \beta_X^2 + \beta_X \beta_Y)(1 + \lambda) + \alpha \sigma_X^2 }{\beta_X^2 (1 + \lambda) + \sigma_X^2}.
\end{equation*}
If $\lambda = -1$, this recovers the causal effect of $X$ on $Y$, and if $\lambda \rightarrow \infty$, this recovers a set of coefficients that are invariant to variation in $A$, as can be seen by plugging the resulting coefficient $\gamma = \alpha  + \beta_Y / \beta_X$ into Equation~\eqref{eq:ex1_worst_case_error}.

\section{Proofs}%
\label{sec:proofs}

\subsection{Auxiliary results}
\label{sub:aux_proofs}

First, we show that the proof of Theorem 1 of \citet{Rothenhausler2018} can be decomposed into two parts, and use this observation to simplify the proof of our Theorem~\ref{thmthm:single_proxy_ar_guarantee}.  Proposition~\ref{thmprop:pop_version_iv_losses} establishes that $\ell_{PLS}$ can be written as a quadratic form in the structural parameters $w_{\gamma}^\top M_A$.  Proposition~\ref{thmprop:general_ar_guarantee} is a straightforward generalization of the techniques used in \citet{Rothenhausler2018}, and establishes that any regularization term that can be written in this way naturally implies a robustness guarantee.

By Assumption~\ref{asmp:linear_scm}, our SCM can be written in the following form, where $\epsilon \indep A$, and all variables are mean-zero and have bounded covariance.
\begin{equation}
  \begin{pmatrix} X \\ Y \\ H \end{pmatrix} = {(Id - B)}^{-1}(M_A A + \epsilon). \label{eq:ar_scm_general}
\end{equation}
In this context, we use the following notational shorthand, 
\begin{equation}
w_\gamma \coloneqq {\left({(Id - B)}^{-1}_{d_X+1, \cdot} - \gamma^\top {(Id - B)}^{-1}_{1:d_X, \cdot}\right)}^\top, \label{eq:w_gamma}
\end{equation}
such that we can write the residual as a function of both the exogenous noise $\epsilon$ and $A$ as 
\begin{equation}
  R(\gamma) \coloneqq Y - \gamma^\top X = w_\gamma^\top (\epsilon + M_A A), \label{eq:residual_linear_in_A}
\end{equation}
under the training distribution.
(This identity explains the valley in the loss landscape displayed in Figure~\ref{fig:scatter-intervention}:
If $d_A \geq 2$, for any parameter $\gamma$, there exist an orthogonal intervention direction $\nu \in (w_{\gamma}^\top M_A)^\perp$, to which the loss is invariant.)

\begin{thmappprop}\label{thmprop:pop_version_iv_losses}
Under Assumption~\ref{asmp:linear_scm},
\begin{align}
  &\ell_{PLS}(X, Y, A; \gamma) \nonumber \\
  &= w_{\gamma}^\top M_A \E[AA^\top ] M_A^\top w_{\gamma},\label{eq:pop_version_iv_losses}
\end{align}
and
\begin{align}
& \ell_{PLS}(X,Y,W;\gamma) \nonumber \\
&= w_{\gamma}^\top M_A \E[AW^\top ] {\E[WW^\top ]}^{-1} \E[W A^\top] M_A^\top w_{\gamma}, \label{eq:OmegaW}
\end{align}
where $w_\gamma$ is defined by Equation~\eqref{eq:w_gamma}.
\end{thmappprop}
\begin{proof}
The first statement follows from Equation~\eqref{eq:def_pls_linear} and the observation that
\begin{align*}
  \E[R(\gamma) A^\top] &= \E[w^\top_{\gamma}(\epsilon + M_A A) A^\top] \\
                       &= w_{\gamma}^\top \E[\epsilon A^\top] + w_{\gamma}^\top M_A \E[AA^\top] \\
                       &= w_{\gamma}^\top M_A \E[AA^\top],
\end{align*}
where we used $\epsilon \indep A$.
Similarly
\begin{align*}
  &\ell_{PLS}(X, Y, W; \gamma) \nonumber \\
  &= \E[R(\gamma) W^\top] {\E[WW^\top]}^{-1} \E[W {R(\gamma)}^\top] \nonumber \\
    &=\E[w^\top _{\gamma}(\epsilon + MA) W^\top ] {\E[WW^\top]}^{-1} \E[W {R(\gamma)}^\top] \nonumber\\
    &=w_{\gamma}^\top M_A \E[AW^\top ] {\E[WW^\top ]}^{-1} \E[W A^\top] M_A^\top w_{\gamma},
  \end{align*}
  where the first equality follows from Equation~\eqref{eq:def_pls_linear}, and the final equality
  follows from the fact that $\epsilon \indep W$.
\end{proof}
\begin{thmappprop}\label{prop:general-loss-worst-case}\label{thmprop:general_ar_guarantee}
Under Assumption~\ref{asmp:linear_scm}, for any $\lambda$ and any real, symmetric $\Omega$ such that $0\preceq \E[AA^\top] + \lambda \Omega$, any loss function of the form 
\begin{align}
  \ell(\gamma, \lambda) := \ell_{LS}(X, Y; \gamma) + \lambda w_{\gamma}^\top M_A \Omega M_A^\top w_{\gamma},
  \label{eq:ar_general_loss}
\end{align}
where $w_\gamma$ is defined by Equation~\eqref{eq:w_gamma}, is equal to the following worst-case loss under bounded perturbations
\begin{align*}
  \ell(\gamma, \lambda) = \sup_{\nu \in C(\lambda)} \E_{do(A \coloneqq \nu)}[{(Y - \gamma^\top X)}^2],
  \end{align*}
where
\begin{equation*}
  C(\lambda) \coloneqq \{ \nu: \E[\nu \nu^\top ] \preceq \E[AA^\top ] + \lambda \Omega \}.
\end{equation*}
\end{thmappprop}
\begin{proof}
  We have, making use of the fact that $\epsilon \indep A$, and $\E[\epsilon] = 0$
    \begingroup 
    \allowdisplaybreaks
  \begin{align*}
    &\sup_{\nu \in C(\lambda)} \E_{do(A \coloneqq \nu)} \left[{(Y - \gamma^\top X)}^2\right] \\
    &= \sup_{\nu \in C(\lambda)} \E_{do(A \coloneqq \nu)}\left[{(w^\top _{\gamma}(\epsilon + {M_A} \nu))}^2\right]  \\
    &= \E\left[{(w^\top _{\gamma}\epsilon)}^2\right] +\sup_{\nu \in C(\lambda)} \E[{(w^\top _{\gamma} {M_A} \nu)}^2]  \\
    &= \E\left[{(w^\top _{\gamma}\epsilon)}^2\right] +\sup_{\nu \in C(\lambda)} w^\top _{\gamma} M_A \E[\nu \nu^\top ] {M_A}^\top w_{\gamma}\\
    &= \E\left[{(w^\top _{\gamma}\epsilon)}^2\right] + w^\top _{\gamma} {M_A} (\E[AA^\top ] + \lambda \Omega) {M_A}^\top w_{\gamma} \\
    &= \E\left[{(w^\top _{\gamma}\epsilon)}^2\right] + w^\top _{\gamma} {M_A} \E[AA^\top ] {M_A}^\top w_{\gamma} \\ &\quad + \lambda w^\top _{\gamma} {M_A} \Omega {M_A}^\top w_{\gamma} \\ 
    &= \E\left[{(w^\top _{\gamma}(\epsilon + {M_A} A))}^2\right] + \lambda w^\top _{\gamma} {M_A} \Omega {M_A}^\top w_{\gamma} \\
    &= \ell_{LS}(X, Y; \gamma) + \lambda w^\top _{\gamma} {M_A} \Omega {M_A}^\top w_{\gamma} \\
    &= \ell(\gamma, \lambda),
  \end{align*}
  \endgroup
  where in the fifth line we used the definition of $C(\lambda)$. The supremum is achievable even if $\nu$ is a deterministic vector, since we can take $\nu \coloneqq \frac{S b}{\sqrt{b^\top S b}}$ where $S \coloneqq \E[AA^\top] + \lambda \Omega$ and $b \coloneqq {M_A}^\top w_{\gamma}$.  Then the supremum value is achieved by $\nu$, as $\nu \nu^\top = \frac{S bb^\top S}{b^\top S b} $ and $b^\top \nu \nu^\top b = \frac{b^\top S b b^\top S b}{b^\top S b} = b^\top S b$. To show that $\nu \nu^\top\preceq S$, such that $\nu\in C(\lambda)$, we can take any conformable vector $x$ to see that 
\begin{align*}
  x^\top (S - \nu \nu^\top) x &= x^\top S x - \frac{x^\top S bb^\top S x}{b^\top S b} \\
                                &= \langle x, x \rangle - \frac{\langle x, b \rangle^2}{\langle b, b \rangle} \\
                                &\geq 0,
\end{align*}
where we use the fact that 
$\langle e,f\rangle \coloneqq  e^\top S f$ defines an inner product, and we apply Cauchy-Schwarz: $\langle x, x \rangle \langle b, b \rangle \geq \langle x, b \rangle^2$.
\end{proof}

In the proofs for Section~\ref{sec:worst_case_bounded_shift}, we will occasionally make use of the following fact, which we prove here to simplify exposition later on.
\begin{thmappprop}\label{thmprop:omega_w_psd_smaller}
In the setting of a single proxy (i.e., under Assumptions~\ref{asmp:linear_scm} and~\ref{asmp:ar_scm_one_proxy}) let $\Omega_W$ be defined as follows
\begin{equation}
  \Omega_W \coloneqq \E[AW^\top ]{\E[WW^\top ]}^{-1}\E[WA^\top ].\label{eq:def_omega_w_app}
\end{equation}
Then $\Omega_W \preceq \E[AA^\top]$. Furthermore, if $\E[\epsilon_W \epsilon_W^\top]$ is positive definite, then this inequality is strict, that is,  $\Omega_W \prec \E[AA^\top]$.
\end{thmappprop}
\begin{proof}
Recall that $\E[AA^\top]$ and $\E[WW^\top]$ are invertible (and hence positive definite) by assumption.  

The inequality $\Omega_W \preceq \E[AA^\top]$ is equivalent to showing that $S\coloneqq  \E[AA^\top] - \E[AW^\top]{\E[WW^\top]}^{-1}\E[WA^\top] \succeq 0$.  Observe that $S$ is the Schur complement of the matrix 
$K \coloneqq  \E\left[\begin{pmatrix}A \\ W\end{pmatrix}\begin{pmatrix}A \\ W\end{pmatrix}^\top\right]$.  The matrix $K$ is positive semi-definite (PSD) if and only if $\E[AA^\top]$ is positive definite (true by assumption) and $S$ is PSD\@ (see \citet[Theorem~1.12b]{zhang2006schur}).  Since $K$ is PSD by construction, as the covariance matrix of $A, W$, this implies that $S \succeq 0$.

Similarly, $K$ is positive definite (PD) if and only if $\E[AA^\top]$ and $S$ are both PD (see \citet[Theorem~1.12a]{zhang2006schur}).  Under the condition that $\E[\epsilon_W \epsilon_W^\top]$ is full-rank, then $K$ is PD, and the second inequality follows.
\end{proof}

\subsection{Proof of additional results}\label{sub:add_proofs}
\begin{proof}[Proof of Equation~\eqref{eq:cov_id}]
    It follows from Proposition~\ref{thmprop:pop_version_iv_losses} that 
    \begin{align*}
        \ell_{PLS}(X,Y,A;\gamma) &= w_\gamma^\top M_A \Omega_A M_A^\top w_\gamma \\
        \ell_{PLS}(X,Y,W;\gamma) &= w_\gamma^\top M_A \Omega_W M_A^\top w_\gamma,
    \end{align*}
    where $\Omega_W := \E[AW^\top]{\E[WW^\top]}^{-1}\E[WA^\top]$ and $\Omega_A := \E[AA^\top]$ are both full rank because $\E[AW^\top] = \E[AA^\top]\beta_W$ and by assumptions that $\E[WW^\top], \E[AA^\top]$ and $\beta_W$ are full rank. Hence both $\ell_{PLS}(X,Y,A;\gamma)$ and $\ell_{PLS}(X,Y,W;\gamma)$ are zero exactly when $w_\gamma^\top M_A = 0$.
\end{proof}

\subsection{Proof of main results}

\subsubsection{Section~\ref{sec:worst_case_bounded_shift}}%
\label{ssub:proofs_section_bounded}
\begin{proof}[Proof of Theorem~\ref{thmthm:single_proxy_ar_guarantee}]
We use the fact that $\epsilon$ is mean-zero and independent of both $A$ and $W$.  Recall that 
\begin{align*}
  \ell_{PAR}(W; \gamma, \lambda) = \ell_{LS}(\gamma) + \lambda \ell_{PLS}(W; \gamma),
\end{align*}
where we suppress the dependence on $X, Y$ in the notation.  Letting $w_\gamma$ be as defined in Equation~\eqref{eq:w_gamma}, it follows from Equation~\eqref{eq:OmegaW} that
\begin{align*}
  &\ell_{PLS}(X, Y, W; \gamma) \nonumber \\
    &=w_{\gamma}^\top M_A \underbrace{\E[AW^\top ] {\E[WW^\top ]}^{-1} \E[W A^\top]}_{\Omega_W} M_A^\top w_{\gamma}.
  \end{align*}
  The statement then follows from the application of Proposition~\ref{thmprop:general_ar_guarantee}, and the fact that $\Omega_W \preceq \E[AA^\top]$ (by Proposition~\ref{thmprop:omega_w_psd_smaller}), such that $\E[AA^\top] + \lambda \Omega_W \succeq 0$ for all $\lambda \geq -1$.
\end{proof}

\begin{proof}[Proof of Proposition~\ref{thmprop:subset_guarantee}]
Recall that the guarantee regions are given by
\begin{align*}
C_{A}(\lambda) &= \{ \nu: \E[\nu \nu^\top ] \preceq \E[AA^\top ] + \lambda \E[AA^\top ] \}  \\
C_{W}(\lambda) &= \{ \nu: \E[\nu \nu^\top ] \preceq \E[AA^\top ] + \lambda \Omega_W \} \\
C_{OLS} &= \{ \nu: \E[\nu \nu^\top ] \preceq \E[AA^\top ] \},
\end{align*}
where \[\Omega_W = \E[AW^\top ]{\E[WW^\top ]}^{-1}\E[WA^\top ]. \]
The fact that $\E{[WW^\top]}^{-1} \succ 0$ implies $\Omega_W \succeq 0$, 
and this implies that $C_{OLS}\subseteq C_W(\lambda)$ for $\lambda \geq 0$.  Showing $C_{W}(\lambda) \subset C_A(\lambda)$ amounts to showing that $\Omega_W \prec \E[AA^\top]$, which holds by Proposition~\ref{thmprop:omega_w_psd_smaller} when $\E[\epsilon_W \epsilon_W^\top] \succ 0$.

Next, we prove that $C_W$ is monotonically decreasing in the noise $\E[\epsilon_W\epsilon_W^\top]$,
in the sense that if $\E [\epsilon_W\epsilon_W^\top] \preceq \E [\eta_W\eta_W^\top]$ then 
\begin{align*}
&\E_\eta[AW^\top] \E_\eta{[WW^\top]}^{-1} \E_\eta[WA^\top]\\ &\preceq \E_\epsilon[AW^\top] \E_\epsilon{[WW^\top]}^{-1} \E_\epsilon[WA^\top],
\end{align*}
where $\E_\eta$ is the expectation in the SCM where $W \coloneqq  \beta_W^\top A + \eta_W$ (and similar for $\E_\epsilon$).

Suppose that $\E[\epsilon_W\epsilon_W^\top] \preceq \E[\eta_W\eta_W^\top]$. Then $\E_\eta{[WW^\top]}^{-1} \preceq \E_\epsilon{[WW^\top]}^{-1}$, and since $\E_\eta[AW^\top] = \E_\epsilon[AW^\top]$, for any vector $x \in \R^{d_A}$ it holds that,
\begin{align*}
    & {(\E_\eta[WA^\top]x)}^\top\E_{\eta}{[WW^\top]}^{-1}(\E_\eta[WA^\top]x) \\
    &\leq {(\E_\epsilon[WA^\top]x)}^\top\E_{\epsilon}{[WW^\top]}^{-1}(\E_\epsilon[WA^\top]x).
\end{align*}
This establishes the matrix inequality. 

To conclude the proof, suppose that $\E[\epsilon_W\epsilon_W^\top] = 0$, $d_A = d_W$ and that $\beta_W$ has full rank. It then follows that
\begin{align*}
  \Omega_W &= \E[AW^\top]\E{[WW^\top]}^{-1}\E[WA^\top] \\
             &= \E[AA^\top]\beta_W{(\beta_W^\top \E[AA^\top]\beta_W)}^{-1}\beta_W^\top \E[AA^\top] \\
             &= \E[AA^\top] \beta_W \beta_W^{-1} {\E[AA^\top]}^{-1} {\beta_W^\top}^{-1}\beta_W^\top \E[AA^\top] \\
    &= \E[AA^\top],
\end{align*}
such that $C_W(\lambda) = \E[AA^\top] + \lambda \Omega_W = (1+\lambda) \E[AA^\top] = C_A(\lambda)$.
\end{proof}
\begin{proof}[Proof of Theorem~\ref{thmthm:cross_equals_iv}]
Let $w_\gamma$ be defined as in Equation~\eqref{eq:w_gamma}.
We can write the population quantity as follows, making use of the fact that $\epsilon$, $\epsilon_Z$, and $\epsilon_W$ are jointly independent, and that all errors have zero mean.
\begingroup
\allowdisplaybreaks
\begin{align*}
    &\ell_{\times}(W, Z; \gamma) \\
    &= {\E[{(Y - \gamma^\top X)} W^\top ]} {\E[ZW^\top ]}^{-1} {\E[Z {(Y - \gamma^\top X)}^\top ]}  \\
    &= \E[w_\gamma^\top (M_A A + \epsilon) W^\top ] {\E[ZW^\top ]}^{-1} \\ &\quad\quad\cdot \E[Z (A^\top M_A^\top + \epsilon^\top )w_{\gamma}] \\
    &= w_{\gamma}^\top M_A \E[AW^\top ] {\E[ZW^\top ]}^{-1} \E[ZA^\top ] M_A^\top w_{\gamma} \\
    &= w_{\gamma}^\top M_A \E[A(A^\top \beta_W + \epsilon_W^\top )] \\ &\quad\quad{\E[(\beta_Z^\top A + \epsilon_Z)(A^\top \beta_W + \epsilon_W^\top )]}^{-1} \\ &\quad\quad\E[(\beta_Z^\top A + \epsilon_Z)A^\top ] M_A^\top w_{\gamma} \\
    &= w_{\gamma}^\top M_A \E[AA^\top ] \beta_W {\left(\beta_Z^\top \E[AA^\top ] \beta_W\right)}^{-1} \\&\quad\quad\beta_Z^\top \E[AA^\top ] M_A^\top w_{\gamma} \\
    &= w_{\gamma}^\top M_A \E[AA^\top ] \beta_W \beta_W^{-1} {\E[AA^\top ]}^{-1} {(\beta_Z^\top )}^{-1} \\&\quad\quad\beta_Z^\top \E[AA^\top ] M_A^\top w_{\gamma} \\
    &= w_{\gamma}^\top M_A \E[AA^\top ]  {\E[AA^\top ]}^{-1}  \E[AA^\top ] M_A^\top w_{\gamma} \\
    &= w_{\gamma}^\top M_A \E[AA^\top ] M_A^\top w_{\gamma}
\end{align*}
\endgroup
The result follows from Proposition~\ref{thmprop:pop_version_iv_losses}.
\end{proof}

In the main text, we state that the xPAR($W,Z$) objective is convex in $\gamma$ and has a closed form solution. We give the proof here:
\begin{thmappprop}\label{thmprop:xpar_closed_form}
Under Assumptions~\ref{asmp:linear_scm},~\ref{asmp:ar_scm_two_proxy} and~\ref{asmp:zw_invert}, the loss in Equation~\eqref{eq:xpar_objective} is convex in $\gamma$, and its minimizer is given by
\begin{align*}
    \gamma_{\times PAR}^* &\coloneqq \left( 2\E[XX^\top] + \lambda (L + L^\top)\right)^{-1} \\
    &\quad\quad \left(2 \E[XY^\top] + \lambda (K_1 + K_2) \right),
\end{align*}
where we define
\begin{align*}
L &:= \E[XW^\top]\E[ZW^\top]^{-1}\E[ZX^\top], \\ 
K_1 &:= \E[XW^\top]\E[ZW^\top]^{-1}\E[ZY^\top] \\
K_2 &:= \E[XZ^\top]\E[WZ^\top]^{-1}\E[WY^\top].
\end{align*}
\end{thmappprop}
\begin{proof}
By Theorem~\ref{thmthm:cross_equals_iv} and Equation~\eqref{eq:arguarantee}, $\ell_{\times PAR}(W, Z;\gamma, \lambda) = \ell_{AR}(X,Y,A;\gamma,\lambda)$, and the latter is convex in $\gamma$, since it is the sum $\ell_{LS}$, which is convex, and $\lambda\ell_{PLS}(X, Y, A;\gamma)$, which is a quadratic form by Proposition~\ref{thmprop:pop_version_iv_losses} and hence convex.

Consequently optimal solution can be found by taking the gradient of $\ell_{\times PAR}(W, Z;\gamma, \lambda) = \ell_{LS} + \lambda \ell_{\times}$ with respect to $\gamma$ and equating it to $0$.  Letting $D := \E[ZW^\top]^{-1}$, we can differentiate $\ell_{\times PAR}$ term wise, using Equation~\eqref{eq:cross_regularization} to rewrite $\ell_\times$:
\begin{align*}
    0 &= 2\gamma^\top \E[XX^\top] - 2\E[YX^\top] \\
    & \quad -\lambda\E[YW^\top] D \E[ZX^\top] \\
    & \quad -\lambda\E[YZ^\top]D^\top \E[WX^\top] \\
    & \quad +\lambda\gamma^\top (L + L^\top),
\end{align*}
where $L := \E[XW^\top]\E[ZW^\top]^{-1}\E[ZX^\top]$. Defining $K_1 := \E[XW^\top]D\E[ZY^\top]$ and $K_2 := \E[XZ^\top]D^\top\E[WY^\top]$, and rearranging, we obtain:
\begin{align*}
    &\gamma^\top(2 \E[XX^\top] + \lambda (L + L^\top)) \\ &= 2 \E[YX^\top] + \lambda (K_1^\top + K_2^\top),
\end{align*}
so by transposing and solving for $\gamma$, we get the expression from the statement. 
\end{proof}

\subsubsection{Section~\ref{sec:targeted_shift}}%
\label{ssub:proofs_section_target}

\begin{proof}[Proof of Proposition~\ref{thmprop:targeting_known_shift}]
Let $w_\gamma$ be defined by \eqref{eq:w_gamma} and for any $\gamma$ let $b_\gamma^\top \coloneqq w_\gamma^\top M_A$. We can write the loss as follows
\begingroup
\allowdisplaybreaks
\begin{align*}
    &\E_{do(A \coloneqq \nu)}[{(Y - \gamma^\top X - \alpha)}^2] \\ 
   &= \E[{(w_{\gamma}^\top (\epsilon + M_A \nu) - \alpha)}^2] \\
   &= \E[{(w_{\gamma}^\top \epsilon + w_{\gamma}^\top M_A \nu - \alpha)}^2] \\
   &\stackrel{\epsilon \indep \nu}{=} \E[{(w_{\gamma}^\top \epsilon)}^2] + \E[{(w_{\gamma}^\top M_A \nu - \alpha)}^2] \\
   &= \E[{(w_{\gamma}^\top \epsilon)}^2] + \E[{(w_{\gamma}^\top M_A A)}^2] \\
     &\quad\quad - \E[{(w_{\gamma}^\top M_A A)}^2] + \E[{(w_{\gamma}^\top M_A \nu - \alpha)}^2] \\
    &= \ell_{LS}(\gamma) - \E[{(b_{\gamma}^\top A)}^2] + \E[{(b_{\gamma}^\top \nu - \alpha)}^2] \\
    &= \ell_{LS}(\gamma) - b_{\gamma}^\top \E[A A^\top] b_{\gamma}^\top \\
     &\quad\quad + b_{\gamma}^\top \E[\nu \nu^\top] b_{\gamma} - 2 \E[b_{\gamma}^\top \nu] \alpha + \alpha^2 \\
    &= \ell_{LS}(\gamma) + b_{\gamma}^\top \left(\E[\nu \nu^\top] - \E[AA^\top]\right) b_{\gamma} \\
     &\quad\quad - 2 \E[b_{\gamma}^\top \nu] \alpha + \alpha^2 \\
    &= \ell_{LS}(\gamma) \\
     &\quad\quad + b_{\gamma}^\top \left(\E[\nu \nu^\top] - \E[AA^\top]\right) b_{\gamma} - {(b_{\gamma}^\top \E[\nu])}^2 \\
    &\quad\quad + {(b_{\gamma}^\top \E[\nu])}^2 - 2 \E[b_{\gamma}^\top \nu] \alpha + \alpha^2 \\
    &= \ell_{LS}(\gamma) + b_{\gamma}^\top \left(\Sigma_{\nu} - \Sigma_{A}\right) b_{\gamma} + {\left(b_{\gamma}^\top \E[\nu]- \alpha \right)}^2,
\end{align*}
\endgroup
where for any value of $\gamma$, that minimizing with respect to $\alpha$ yields $\alpha^* = b_{\gamma}^\top \E[\nu]$, where $b_{\gamma}^\top = w_{\gamma}^\top M_A$.  Given that we can write the structural relationship $Y - \gamma^\top X = b_{\gamma}^\top A + w_{\gamma}^\top \epsilon$, and knowing that $\E[\epsilon] = 0$ and that $\epsilon \indep A$, we know that $b_{\gamma}^\top A$ is the conditional expectation of $R(\gamma)$ given $A$.
\end{proof}

In the main text, we note that Equation~\eqref{eq:tar_a} (the objective function $\ell_{TAR}$) is convex in $\gamma, \alpha$, and has a closed form solution.  We prove that result here.
\begin{thmappprop}\label{thmprop:convexity_known_shifts}
Under Assumption~\ref{asmp:linear_scm}, the minimizer $\gamma^*_{TAR}, \alpha^*_{TAR}$ of Equation~\eqref{eq:tar_a} is given by 
\begin{align*}
  \gamma^* &= {\left(\E[XX^\top] + \E[XA^\top] \Omega \E[AX^\top]\right)}^{-1}\\ & \quad\left(\E[XY^\top] + \E[XA^\top] \Omega \E[AY^\top]\right) \\
  \alpha^* &= b_{\gamma^*}^\top \mu_{\nu},
\end{align*}
where $\Omega = {\E[AA^\top]}^{-1} (\Sigma_\nu - \Sigma_A) {\E[AA^\top]}^{-1}$, and $b_{\gamma}^\top$ is defined in Equation~\eqref{eq:def_b_gamma}.
\end{thmappprop}
\begin{proof}[Proof of Proposition~\ref{thmprop:convexity_known_shifts}]
Let $w_\gamma$ be as defined in Equation~\eqref{eq:w_gamma} and let $b_\gamma^\top \coloneqq  w_\gamma^\top M_A$.
Since $\E[(Y - \gamma^\top X)\mid A] = \E[w_\gamma^\top[M_A A + \epsilon]\mid A] = b_\gamma^\top A$,  for any $\gamma$, $b_\gamma^\top$ is the linear regression coefficient of $(Y - \gamma^\top X)$ onto $A$, so we may write $b_\gamma^\top = \E[(Y- \gamma^\top X)A^\top]\E[AA^\top]^{-1}$. Plugging in the optimal value $\alpha(\gamma) \coloneqq b_\gamma^\top \mu_\nu$, we obtain 
    \begin{align*}
        &\ell_{TAR}(A; \mu_{\nu}, \Sigma_{\nu}, \gamma, \alpha(\gamma)) \\
        &= \ell_{LS}(\gamma) + b_{\gamma}^\top \left(\Sigma_{\nu} - \Sigma_{A}\right) b_{\gamma} \\
        &= \ell_{LS}(\gamma) + \E[(Y- \gamma^\top X)A^\top] \Omega \E[A {(Y- \gamma^\top X)}^\top]
    \end{align*}
This objective is convex in $\gamma$. 
The derivative of the loss with respect to $\gamma$ is
\begin{align*}
    -2 (\E[(Y-\gamma^\top X)X^\top] + \E[(Y - \gamma^\top X) A^\top]\Omega\E[AX^\top]),
\end{align*}
and equating to $0$ and solving for $\gamma$ yields
\begin{align*}
  \gamma^* &= {\left(\E[XX^\top] + \E[XA^\top]\Omega \E[AX^\top]\right)}^{-1}\\
    &\quad\left(\E[XY^\top] + \E[XA^\top] \Omega \E[AY^\top]\right).
\end{align*}
\end{proof}

We also claim in the main text that if $\nu$ is a constant, then the minimizer of Equation~\eqref{eq:tar_a} can be found by performing OLS using both $X, A$ as predictors, and then plugging in the known value $\nu$ for $A$ in prediction.  We prove that result here.
\begin{proof}
If $\nu$ is a constant, then we can write the first two terms as follows, where $w_{\gamma}$ is defined in Equation~\eqref{eq:w_gamma}.
\begin{align*}
  &\ell_{LS} - b_{\gamma}^\top \Sigma_A b_{\gamma} \\
  &= \E[{(w_{\gamma}^\top (M_A A + \epsilon))}^2] - w_{\gamma}^\top M_A \E[AA^\top] M_A^\top b_{\gamma} \\
  &= \E[{(w_{\gamma}^\top (M_A A + \epsilon))}^2] - \E[{(w_{\gamma}^\top M_A A)}^2] \\
  &= \E[{(w_{\gamma}^\top \epsilon)}^2] 
\end{align*}
which is equivalent to the objective for the loss when $Y, X$ are residualized with respect to $A$ (see Section 8.6 of \citet{Rothenhausler2018}).  By the Frish-Waugh-Lovell theorem \citep{Lovell1963, Lovell2008}, this yields the same coefficients $\gamma$ for $X$ as if we had performed regression on $X, A$ together.  For this value of $\gamma$, $b_{\gamma}^\top$ is the coefficient that we would obtain for $A$ in the joint regression, because it equals the regression coefficients for $Y - \gamma^\top X$ on $A$.
\end{proof}

\begin{proof}[Proof of Proposition~\ref{thmprop:generalized_ar}]
We use $\nu$ to denote the random shift. Let $\nu \in T(\mu_{\nu}, \Sigma_\nu)$, or equivalently, let $\nu \coloneqq \mu_\nu + \delta$, where $\mu_\nu$ is fixed and $\delta$ satisfies the constraint that $\E[\delta \delta^\top] \preceq \Sigma_\nu$,  where $\Sigma_{\nu}$ is a symmetric positive definite matrix.  Let $w_\gamma$ be defined by~\eqref{eq:w_gamma} and for any $\gamma$ let $b_\gamma^\top \coloneqq w_\gamma^\top M_A$. We can write the loss as follows
\begin{align*}
    &\sup_{\nu \in T} \E_{do(A \coloneqq \nu)}[{(Y - \gamma^\top X - \alpha)}^2] \\ 
   &=\sup_{\nu \in T} \E[{(w_{\gamma}^\top (\epsilon + M_A \nu) - \alpha)}^2] \\
   &= \sup_{\nu \in T} \E[{(w_{\gamma}^\top \epsilon + w_{\gamma}^\top M_A \nu - \alpha)}^2] \\
   &= \E[{(w_{\gamma}^\top \epsilon)}^2] + \sup_{\nu \in T} \E[{(w_{\gamma}^\top M_A \nu - \alpha)}^2] \\ 
   &= \E[{(w_{\gamma}^\top \epsilon)}^2] + \E[{(w_{\gamma}^\top M_A A)}^2] \\
   &\quad\quad - \E[{(w_{\gamma}^\top M_A A)}^2] + \sup_{\nu \in T} \E[{(w_{\gamma}^\top M_A \nu - \alpha)}^2] \\
   &= \ell_{LS}(\gamma) - \E[{(b_{\gamma}^\top A)}^2] + \sup_{\nu \in T} \E[{(b_{\gamma}^\top \nu - \alpha)}^2],
\end{align*}
where on the fourth line we used the fact that $\E[\epsilon \nu] = 0$ by the fact that $\nu = \mu_v + \delta$, and $\delta$ is independent of $\epsilon$.  In the last line we replaced $w_{\gamma}^\top M_A$ by $b_{\gamma}^\top$.  We can re-write the last term as follows, where the supremum with respect to $\delta$ is constrained in the set $\E[\delta \delta^\top] \preceq \Sigma_\nu$
\begin{align*}
 &\sup_{\nu \in T} \E[{(b_{\gamma}^\top \nu - \alpha)}^2] \\
 &=\sup_{\delta: \E[\delta \delta^\top] \preceq \Sigma_{\nu}} \E[{(b_{\gamma}^\top (\delta + \mu_{\nu}) - \alpha)}^2] \\
 &=\sup_{\delta} \E[{(b_{\gamma}^\top \delta + b_{\gamma}^\top \mu_{\nu} - \alpha)}^2] \\
 &=\sup_{\delta} \E[{(b_{\gamma}^\top \delta)}^2] + 2 \E[(b_{\gamma}^\top \delta)]( b_{\gamma}^\top \mu_{\nu} - \alpha) + \E[{(b_{\gamma}^\top \mu_{\nu} - \alpha)}^2] \\
 &=b_{\gamma}^\top \Sigma_{\nu} b_{\gamma} + 2 \norm{b_{\gamma}}_{\Sigma_\nu} \cdot \abs{ b_{\gamma}^\top \mu_{\nu} - \alpha} + {(b_{\gamma}^\top \mu_{\nu} - \alpha)}^2,
\end{align*}
where $\norm{b_{\gamma}}_{\Sigma_\nu} \coloneqq \sqrt{b_{\gamma}^\top \Sigma_\nu b_{\gamma}}$ is the norm induced by the inner product defined with respect to $\Sigma_{\nu}$.  In the last line, we have used the fact that the expression is maximized (subject to the constraint) by the deterministic distribution $\delta_* = \pm \frac{\Sigma_{\nu} b_{\gamma}}{\sqrt{b_{\gamma}^\top \Sigma_{\nu} b_{\gamma}}}$ where the sign depends on the sign of $(b_{\gamma}^\top \mu_{\nu} - \alpha)$: $\delta_*$ satisfies $b_\gamma^\top \delta_*\delta_*^\top b_\gamma = b_\gamma^\top \Sigma_\nu b_\gamma$, 
maximizing the first term. Further, the second term is also maximized by $\delta_*$, because 
if any other random or deterministic $\delta$ satisfies $|\E b_\gamma^\top \delta| > |b_\gamma^\top \delta_*|$, it follows by Jensens inequality that $\E[(b_\gamma^\top \delta)^2] \geq (\E[(b_\gamma^\top \delta)])^2 > (b_\gamma^\top \delta_*)^2 = b_\gamma^\top \Sigma_\nu b_\gamma$, such that $\E[\delta\delta^\top] \succ \Sigma_\nu$, so $\delta$ is not in the set over which the supremum is taken. Consequently, the supremum is attained at $\delta_*$, because  $\delta_*$ maximizes both terms.

Using this expression for the supremum, we can write the objective as
\begin{align*}
  &\sup_{\nu \in T} \E_{do(A \coloneqq \nu)}[{(Y - \gamma^\top X - \alpha)}^2] \\ 
  &= \ell_{LS}(\gamma) + b_{\gamma}^\top(\Sigma_{\nu} - \Sigma_A) b_{\gamma} \\
  &\quad\quad + 2 \norm{b_{\gamma}}_{\Sigma} \cdot \abs{b_{\gamma}^\top \mu_{\nu} - \alpha} + {(b_{\gamma}^\top \mu_{\nu} - \alpha)}^2,
\end{align*}
for which the optimal choice of $\alpha^*$ is given by $b_{\gamma}^\top \mu_{\nu}$, for any $\gamma$, and for this choice of $\alpha$, we can see that $\gamma^* = \argmin_{\gamma} \ell_{LS}(\gamma) + b_{\gamma}^\top \left( \Sigma_{\nu} - \Sigma_{A} \right) b_{\gamma}$.
\end{proof}

\section{Targeting with proxies}
\label{sub:targeted_shift_one_proxy}

\begin{thmdef}[Proxy Targeted Anchor Regression]\label{def:ptar_objective}
Let $\tilde{\mu} \coloneqq \E_{do(A \coloneqq \nu)}[W]$ denote the mean of $W$ under intervention, and let  $\tilde{\Sigma}_{W} \coloneqq \cov_{do(A \coloneqq \nu)}(W)$ denote the covariance. We define
\begin{align}
  &\ell_{PTAR}(W; \tilde{\mu}, \tilde{\Sigma}_{W}, \gamma, \alpha)\label{eq:tar_w} \\
  &= \ell_{LS}(\gamma) + c_{\gamma}^\top \left(\tilde{\Sigma}_{W} - \Sigma_W\right) c_{\gamma} + {(c_{\gamma}^\top \tilde{\mu} - \alpha )}^2, \nonumber
\end{align}
where $c_{\gamma}^\top \coloneqq \E[R(\gamma) W^\top] {\Sigma_W}^{-1}$. 
\end{thmdef}
As mentioned in the main text, Equation~\eqref{eq:tar_w} is not generally equal to Equation~\eqref{eq:tar_a}, and does not generally yield the optimal predictor under the targeted loss.  A simple example is given in Proposition~\ref{thmprop:ptar_underestimates}.
\begin{thmappprop}\label{thmprop:ptar_underestimates}
Assume Assumptions~\ref{asmp:linear_scm},~\ref{asmp:ar_scm_one_proxy}, and that $\E[\epsilon_W \epsilon_W^\top]$ is full rank. Let $\nu \eid A + \eta$ for the deterministic vector $\eta^T = \E[R(\gamma^*_{OLS}) A^\top]$, where $\eid$ indicates equality of distribution, and assume $\eta \neq 0$.  Then, the minimizers of Equations~\eqref{eq:tar_a} and~\eqref{eq:tar_w} differ, in that
\begin{equation*}
  \alpha^*_{PTAR} < \alpha^*_{TAR}
\end{equation*}
and if $d_W = d_A = 1$, and $A$ has unit variance, then $\frac{\alpha^*_{PTAR}}{\alpha^*_{TAR}} = \rho_W$, where $\rho_W \coloneqq \beta_W^2 / (\beta_W^2 + \E[\epsilon_W^2])$.
\end{thmappprop}
\begin{proof}
The assumption that $\nu = A + \eta$ implies that $\Sigma_\nu - \Sigma_A = 0$, and $\E[\nu] = \eta$.  That is, we have changed the mean of the distribution, but not the covariance.  This implies
\begin{align*}
  \E[\tilde{W}] = \beta_W^\top \E[\nu] &= \beta_W^\top \eta \\
  \Sigma_{\tilde{W}} - \Sigma_{W} = \beta_W^\top (\Sigma_{\nu} - \Sigma_{A}) \beta_W &= 0,
\end{align*}
where in the second equation we use the fact that $\Sigma_{W} = \beta_W^\top \E[AA^\top] \beta_W + \E[\epsilon_W \epsilon_W^\top]$ (and similarly for $\Sigma_{\tilde{W}}$), and the $\epsilon_W$ terms cancel in the subtraction.  We can then write both objectives as follows
\begin{align*}
  &\ell_{PTAR}(W, \tilde{W}; \gamma, \alpha) \\
  & \quad \quad = \ell_{LS}(\gamma) + {\left( c_{\gamma}^\top \beta_W^\top \eta - \alpha  \right)}^2 \\
                                            &\quad \quad = \ell_{LS}(\gamma) + {\left(  \E[R(\gamma) A^T] \beta_W \Sigma_{W}^{-1} \beta_W^\top \eta - \alpha  \right)}^2 \\
                                            &\ell_{TAR}(A, \nu; \gamma, \alpha) \\
                                            & \quad \quad = \ell_{LS}(\gamma) + {\left( b_{\gamma}^\top \eta - \alpha  \right)}^2 \\
                                     & \quad \quad = \ell_{LS}(\gamma) + {\left(  \E[R(\gamma) A^T] \Sigma_A^{-1} \eta - \alpha  \right)}^2 
\end{align*}
This gives the optimal value of $\alpha$ in both cases as the value that minimizes the second term
\begin{align*}
  \alpha^*_{PTAR} &= \E[R(\gamma^*_{PTAR}) A^T] (\beta_W \Sigma_{W}^{-1} \beta_W^\top ) \eta \\
  \alpha^*_{TAR} &= \E[R(\gamma^*_{TAR}) A^T] \Sigma_A^{-1} \eta,
\end{align*}
and since the second term can be made equal to zero by these choices of $\alpha$, the optimal $\gamma$ in both cases is identically $\gamma^*_{PTAR} = \gamma^*_{TAR} = \gamma^*_{OLS}$, the value of $\gamma$ that minimizes the first term $\ell_{LS}(\gamma)$.  Hence, we can write the difference between these terms as 
\begin{align*}
  &\alpha^*_{TAR} - \alpha^*_{PTAR} \\
  &= \E[R(\gamma^*_{OLS}) A^T] (\Sigma_A^{-1} - \beta_W \Sigma_{W}^{-1} \beta_W^\top) \E[A R(\gamma^*_{OLS})],
\end{align*}
where we have replaced $\eta$ with the assumed value of $\E[A R(\gamma^*_{OLS})]$.  By assumption, $\Sigma_A$ is full-rank, so that matrix $\Omega \coloneqq (\Sigma_A^{-1} - \beta_W \Sigma_{W}^{-1} \beta_W^\top)$ is positive definite if and only if $\Sigma_A \Omega \Sigma_A$ is positive definite. 
Working with this representation, we can see that 
\begin{align*}
  \Sigma_A \Omega \Sigma_A &= \Sigma_A - \Sigma_A \beta_W \Sigma_{W}^{-1} \beta_W^T \Sigma_A \\
                           &= \E[AA^\top] - \E[AW^\top] {\E[WW^{\top}]}^{-1} \E[WA^\top] \\
                           &\succ 0,
\end{align*}
where the last line follows from Proposition~\ref{thmprop:omega_w_psd_smaller}.  In the case where $d_W = d_A = 1$, and $A$ has unit variance, then let $\rho_W = \beta_W^2 / (\beta_W^2 + \E[\epsilon_W^2])$, and observe that
\begin{align*}
  \alpha^*_{PTAR} &= \eta^2 \rho_W & \alpha^*_{TAR} &= \eta^2.
\end{align*}
\end{proof}

Proposition~\ref{thmprop:ptar_underestimates} describes a worst-case mean-shift in $A$, where $\eta$ is taken in the direction that maximizes the loss of the OLS solution $\gamma^*_{OLS}$.  This is also a particularly simple case to analyze for building intuition, because the optimal solution to both Equations~\eqref{eq:tar_a} and~\eqref{eq:tar_w} is to take $\gamma = \gamma^*_{OLS}$ and to estimate an intercept term $\alpha$ equal to the bias incurred by the shift in the mean of $A$.  However, the noise in $W$ results in under-estimating the impact of the shift, and the gap to the optimal solution depends on the signal-to-variance relationship in $W$, which (as discussed in Section~\ref{sec:worst_case_bounded_shift}) is not generally identified.

We also prove that the Cross-Proxy Targeted Anchor Regression objective is equal to that of Targeted Anchor Regression.

\begin{thmthm}\label{thmthm:two_proxies_target}
  Under Assumptions~\ref{asmp:linear_scm},~\ref{asmp:ar_scm_two_proxy}, and~\ref{asmp:zw_invert}, for all $\gamma \in \R^{d_X}, \alpha \in \R$,
\begin{equation*}
  \ell_{\times TAR}(W, Z; \tilde{\mu}, \tilde{\Sigma}_{W}, \gamma, \alpha) = \E_{do(A\coloneqq\nu)}[{(Y - \gamma^\top X - \alpha)}^2]
\end{equation*}
where $\tilde{\mu} \coloneqq \E_{do(A \coloneqq \nu)}[W]$ is the mean of $W$ under intervention, and $\tilde{\Sigma}_{W}$ is the covariance $\tilde{\Sigma}_{W} \coloneqq \cov_{do(A \coloneqq \nu)}(W)$.
\end{thmthm}
\begin{proof}[Proof of Theorem~\ref{thmthm:two_proxies_target}]
We have
\begin{align*}
  a_{\gamma}^{\top} &= \E[R(\gamma) Z^\top ] {(\E[WZ^\top ])}^{-1}\\
                    &= \E[R(\gamma) (A^\top \beta_Z + \epsilon_Z^\top)]\\ 
                    & \quad\quad {\E[(\beta_W^\top A + \epsilon_W) {(\beta_Z^\top A + \epsilon_Z)}^\top]}^{-1} \\
                         &= \E[R(\gamma) A^\top] \beta_Z {(\beta_W^\top \E[AA^\top] \beta_Z)}^{-1} \\
                         &= \E[R(\gamma) A^\top] {(\E[AA^\top])}^{-1} {(\beta_W^{\top})}^{-1},
\end{align*}
while
\begin{align*}
  \tilde{\mu} &= \beta_W^\top \E[\nu]\\
  \tilde{\Sigma}_{W} - \Sigma_{W} &= \beta_W^\top (\Sigma_{\nu} - \Sigma_{A}) \beta_W. 
\end{align*}
With $b_\gamma^\top \coloneqq  w_\gamma^\top M_A$ and $w_\gamma$ defined by \eqref{eq:w_gamma}, we have that
\begin{align*}
  a_{\gamma}^\top \tilde{\mu} &= b_{\gamma}^\top \E[\nu] \\
  a_{\gamma}^\top (\tilde{\Sigma}_W - \Sigma_W) a_{\gamma} &= b_{\gamma}^\top (\Sigma_\nu - \Sigma_A) b_{\gamma},
\end{align*}
which is equivalent to $\ell_{TAR}(A; \mu_{\nu}, \Sigma_{\nu}, \gamma, \alpha)$ (Definition~\ref{thmdef:tar}, Equation~\eqref{eq:tar_a}).  The proof is complete by Proposition~\ref{thmprop:targeting_known_shift}.
\end{proof}

Note that the argument is symmetric for using an observed shift in either $Z$ or $W$, so it suffices to know the anticipated shift with respect to one proxy. 

\section{Details for experiments}\label{sec:synth_details}
\subsection{Details of Section~\ref{sub:exp-robustness-guarantee}}
\label{sub:details-exp-robustness-guarantee}
We outline the details of the simulation experiment in Section~\ref{sub:exp-robustness-guarantee}.

\paragraph{Summary}
We simulate a training data set $\mathcal{D}_{\text{train}}$ from a SCM that induces the structure in Figure~\ref{fig:causal_graphs}, fix $\lambda \coloneqq  5$ and fit estimators PAR($W$) and xPAR($W, Z$).
We consider the intervention $\P_{do(A\coloneqq \nu)}$ with 
$\nu = (-2.83,0.35,0.71)^\top$, and simulate a test data set $\mathcal{D}_{\text{test}}$ from that distribution.
We then compute the intervention mean squared prediction error (MSPE) 
$\hat{\E}_{do(A\coloneqq \nu)}[(Y - \gamma^\top X)^2]$ both for PAR($W$) and xPAR($W, Z$). We repeat this procedure $m=10^5$ times for several signal-to-variance ratios $x$ (not including $0$), and display the quantiles of the losses in Figure~\ref{fig:loss-robustness}. We also plot the population losses ${\E}_{do(A\coloneqq \nu)}[(Y - \gamma^\top X)^2]$ for PAR($W$) and xPAR($W,Z$), as well as AR($A$) and OLS. 

\paragraph{Technical details}
We let $\E[AA^\top] = \beta = \operatorname{Id}$ and $\E[\epsilon_W\epsilon_W^\top] = s^2 \operatorname{Id}$, such that $W = \beta^\top A + s\cdot \epsilon_W$. Then $\Omega_W$ as defined in Equation~\eqref{eq:single_proxy_ar_guarantee} simplifies to 
\begin{align*}
\Omega_W &=\E[AA^\top]\beta(\beta^\top\E[AA^\top]\beta + \E[\epsilon_W\epsilon_W^\top])^{-1}\beta^\top \E[AA^\top] \\
&=\quad \frac{1}{1 + s^2}\operatorname{Id}.
\end{align*}
We call $x = (1+s^2)^{-1}$ the signal-to-variance ratio, and we can obtain a given signal-to-variance ratio $x$, by setting $s = \sqrt{(1-x)/x}$. 

For each $n \in \{150, 500\}$ and signal-to-variance ratio $x \in \{1/20, 2/20, \ldots, 20/20\}$, we set $s = \sqrt{(1-x)/x}$ and sample a data set $\mathcal{D}^i_{n,s}$ for $i = 1, \ldots, 5000$, each with sample size $n$, from the structural equations:
\begin{align}\label{eq:exp-robustness-structural-eq}
    A &:= \epsilon_A\\
    W &:= A + s\cdot\epsilon_W \nonumber\\
    Z &:= A + s\cdot\epsilon_Z \nonumber\\
    (Y, X, H) &:= (\operatorname{Id}-B)^{-1}(M A + \epsilon), \nonumber
\end{align}
where $d_A = d_W = d_Z = d_X = 3$, $d_Y = d_H = 1$. $M$ and $B$ are given by
\begin{align*}
    M = 
    \begin{pmatrix} 
    1 & 0 & -2 \\ 
    0 & 2 & 1\\
    -1& 3& 0\\
    2& 2& -3\\
    0& -2& 2
    \end{pmatrix}, 
    B = \begin{pmatrix}
    0& -2&  2&  0&  1\\
    0&  0&  0&  0&  0\\
    0&  0&  0&  0&  0\\
    3&  0&  0&  0&  1\\
    0&  0&  0&  0&  0
    \end{pmatrix}, 
\end{align*}
and all noise variables are i.i.d., $\epsilon_A, \epsilon_W, \epsilon_Z, \epsilon \sim \mathcal{N}(0, \operatorname{Id})$. 
For every combination $(n,s)$ we have $5000$ data sets $\mathcal{D}^i_{n,s}$ , $i = 1, \ldots, 5000$. For each data set, we compute the proxy estimators $\gamma_{n,s, W}^{i}$ and $\gamma_{n,s, W; Z}^{i}$, using one or two proxies respectively, and we
simulate $5000$ corresponding test data sets of size $n$ from $\P_{do(A:=\nu)}$ (using the structural equations above, except for changing the assignment for $A$ to $A \coloneqq \nu$). The prediction MSE for the i'th test data set is then $\frac{1}{n}\sum_{j=1}^n (Y_j - \gamma^\top X_j)^2$, resulting in $5000$ values of the MSE for each combination of $(n,s)$.

At each combination of $(n,s)$ we plot the median by a line of the estimated worst case losses, and by a shaded region indicate the interval between the 25\% and 75\% quantiles of the observed distribution. We plot the median instead of the mean since for small $x$, $s^2 = \frac{1-x}{x}$ is large, and especially for $\text{WCL}_{n,s}^i(W, Z)$ and $n = 150$, the mean will be driven very much by outliers for small $x$.

The population versions of losses for any $s$ is computed first by computing the population estimators $\gamma$ from the parameter matrices $M, B$, and then computing the loss at $\nu$ by $\E_{do(A\coloneqq \nu)}[(Y - \gamma^\top X)^2] = w_\gamma^\top M \nu\nu^\top M^\top w_\gamma + w_\gamma^\top \E[\epsilon\epsilon^\top] w_\gamma$. 

\subsection{Details of Section~\ref{sub:exp-misspecified-svr}}
We outline the details of the simulation experiment in Section~\ref{sub:exp-misspecified-svr}. 

\paragraph{Summary}
We analyze the effect of applying anchor regression with one proxy, PAR($W$), when the signal-to-variance ratio is potentially misspecified. 
To do so, we simulate data from the same SCM as in Section~\ref{sub:exp-robustness-guarantee} ($n = 10^4$), and in particular from a range of true (unknown) signal-to-variance ratios $x \in (0, 1]$. To each data set, we apply anchor regression with one proxy, PAR($W$), and with $\lambda:=5$.
We further assume the signal-to-variance ratio to be $40\%$ -- independently of its true value. 
This means, by Proposition~\ref{thmthm:single_proxy_ar_guarantee}, that we assume that PAR($W$) minimizes the worst case mean squared prediction error (MSPE) over the region $C := \{\nu\nu^\top \preceq (1 + 0.4\cdot\lambda) \E[AA^\top]\}$, with the worst case MSPE for being equal to the optimal value of the PAR($W$) objective function.
If $x = 0.4$, then PAR($W$) indeed minimizes the worst case MSPE over $C$ and the estimated worst case MSPE over $C$ is close to the actual worst case MSPE over $C$. But if $x \neq 0.4$, 
the estimator minimizes the worst case MSPE over a different set, and then expect that the true worst case MSPE over $C$ differs from its estimate. 
Figure~\ref{fig:misspecified} shows that this is indeed the case:
We observe that if the true signal-to-variance ratio is larger than the assumed 40\%, our estimate of the MSPE is too conservative. On the contrary, if the true signal-to-variance ratio is smaller than assumed, our estimates of the MSPE over C are too small, meaning that we underestimate the worst case MSPE in the region $C$.

\paragraph{Technical details}

For a fixed signal-to-variance ratio $x$, we simulate a training data set $\mathcal{D}_{\text{train}}$ ($n = 10^4$) from the same procedure as in Section~\ref{sub:details-exp-robustness-guarantee}, i.e. using the structural equations in Equation~\eqref{eq:exp-robustness-structural-eq}, and with the same parameters $M$ and $B$. 
We fit the PAR($W$) estimator to the data using $\lambda := 5$, and the estimated worst case mean squared prediction error (MSPE) over C is then the value of the objective function in the estimated parameter (by Theorem~\ref{thmthm:single_proxy_ar_guarantee}). 

To find the actual worst case MSPE over C for a given estimator $\lambda$, we use the fact from Equation~\eqref{eq:residual_linear_in_A} that 
\begin{align}
\E_{do(A:=v)}[(R - \gamma^\top X)^2] = (b_\gamma^\top v)^2 + w_\gamma^\top w_\gamma \label{eq:exp-misspecified-wcl},
\end{align}
where we use that $\E[\epsilon\epsilon^\top] = \operatorname{Id}$, $w_\gamma$ is given by Equation~\eqref{eq:w_gamma} and $b_\gamma^\top = w_\gamma^\top M_A$. The second term doesn't depend on $v$, and since $C$ is spherical, the worst case MSPE over $C$ is attained in the direction $v\propto b_\gamma$, with $v$ normalized such that $\|v\|^2 = (1+0.4\cdot\lambda)$ (that is $v$ lies on the boundary of $C$). Using the known $M$ and $B$, we compute $w_\gamma, b_\gamma$, and the actual worst case MSPE over $C$ is given by Equation~\eqref{eq:exp-misspecified-wcl} plugging in $v = b_\gamma \cdot \sqrt{(1+0.4\cdot\lambda)}/\|b_\gamma\|$. 

We compute also the worst case MSPE over $C$ when using an OLS estimator for the prediction. We fit $\hat\gamma_{OLS}$ from $\mathcal{D}_{train}$, and, as for the actual MSPE of PAR($W$), the worst case MSPE over $C$ using OLS can be computed, by computing vectors $b_{\hat\gamma_{OLS}}, w_{\hat\gamma_{OLS}}$. Again the worst case MSPE over $C$ using $\hat\gamma_{OLS}$ is attained by setting $v = b_{\hat\gamma_{OLS}} \cdot \sqrt{(1+0.4\cdot\lambda)}/\|b_{\hat\gamma_{OLS}}\|$ and plugging $v$, $b_{\hat\gamma_{OLS}}$ and $w_{\hat\gamma_{OLS}}$ into Equation~\eqref{eq:exp-misspecified-wcl}. 

For every signal-to-variance ratio $x \in \{1/20, \ldots, 20/20\}$, we repeat the procedure $m = 1000$ times, for each computing the estimated and actual MSPEs. In Figure~\ref{fig:misspecified} we plot the median MSPE as well as the interval from the 25\% quantile to the 75\% quantile.

\subsection{Details of Section~\ref{sub:exp-anticausal-causal}}
\label{sub:details-exp-anticausal-causal}
We outline the details of the simulation experiment in Section~\ref{sub:exp-anticausal-causal}. 
\paragraph{Summary}
We demonstrate the ability of Proxy Anchor Regression to select invariant predictors, in a synthetic setting where predictors $X$ may contain both causal and anti-causal predictors.  We simulate data sets ($n=10^5$) from a SCM with the structure shown
in Figure~\ref{fig:scm} (top), where one anchor, $A_1$, is a parent of the causal predictors, while the other $A_2$ is a parent of the anti-causal predictors. 

We consider two identically distributed noisy proxies $W, Z$ of $A\coloneqq (A_1, A_2)$.  The challenge, in this scenario, is that $A_2$ is measured with significantly more noise than $A_1$, across both proxies.
As a consequence, proxy anchor regression with one proxy, PAR($W$), puts more weight on anti-causal features: the noise in $W$ is mistaken for fluctuations in $A_2$, resulting in $X_{\text{anti-causal}}$ mistakenly appearing invariant to shifts in $A_2$. 
In contrast, when two proxies $W, Z$ are available, the estimator xPAR$(W,Z)$ asymptotically equals that of anchor regression with observed anchors, and its regression coefficients puts more weight on the causal predictors; see Figure~\ref{fig:scm} (bottom).

\paragraph{Technical details}

With $d_{A_1} = d_{A_2} = d_{W} = d_{W} = 6$, $d_{X_{\text{causal}}} = d_{X_{\text{anti-causal}}} = 3$ and $d_Y = 1$, we simulate data from the SCM in Figure~\ref{fig:scm} (top) which amounts to simulating from the following structural equations:
\begin{align*}
    A_1 &:= \epsilon_{A_1} \\
    A_2 &:= \epsilon_{A_2} \\
    W &:= (A_1, A_2)^\top + (\epsilon_{W,1}, \epsilon_{W,2})^\top\\
    Z &:= (A_1, A_2)^\top + (\epsilon_{Z,1}, \epsilon_{Z,2})^\top \\
    X_{\text{causal}} &:= M_1 A_1 + \epsilon_{X_{\text{causal}}}\\
    Y &:= \gamma_{\text{causal}}^\top X_{\text{causal}} + \epsilon_{Y}\\
    X_2 &:= M_2 A_2 + \gamma_{\text{anti-causal}} Y + \epsilon_{X_{\text{anti-causal}}}.
\end{align*}
Here $M_1 \in \R^{d_{X_{\text{causal}}} \times d_{A_1}}$ and $M_2 \in \R^{d_{X_{\text{anti-causal}}} \times d_{A_2}}$ are matrices with $1$ in every entry, $\gamma_{\text{causal}} = (1/4, 1/4, 1/4)^\top$ and $\gamma_{\text{anti-causal}} = (4, 4, 4)^\top$ (such that the regression coefficients of $Y$ onto $X_{\text{causal}}, X_{\text{anti-causal}}$ are of similar magnitudes). All noise terms are independent and $\epsilon_{A_1}, \epsilon_{A_2}, \epsilon_{X_{\text{causal}}}, \epsilon_{X_{\text{anti-causal}}}, \epsilon_Y \sim \mathcal{N}(0, \operatorname{Id})$, and $\epsilon_{W,1}, \epsilon_{Z,1} \sim \mathcal{N}(0, \operatorname{Id})$, $\epsilon_{W,2}, \epsilon_{Z,2} \sim \mathcal{N}(0, 3^2\cdot\operatorname{Id})$. 

We simulate a data set $\mathcal{D}$ ($n = 10^5$) from these structural equations, and fit the proxy anchor regression estimators $\gamma(W)$ and $\gamma(W, Z)$ from Section~\ref{sec:worst_case_bounded_shift}.
We repeat this $m=10^4$ times, and display the mean absolute value of the regression coefficients (that is the entries of the vectors $\gamma(W)$ and $\gamma(W, Z)$) in Figure~\ref{fig:scm} (bottom), as well as the standard deviation of the absolute value of the regression coefficients as error bars. 

\subsection{Details of Section~\ref{sub:exp-target-shift}}

\paragraph{Summary}
We demonstrate the trade-off made by Targeted Anchor Regression (TAR) versus Anchor Regression (AR), considering the case when $A$ is observed for simplicity.  We simulate training data and fit estimators $\gamma_{\text{OLS}}$, $\gamma_{\text{AR}}$ and $\gamma_{\text{TAR}}$, where $\gamma_{\text{TAR}}$ is targeted to a particular mean and covariance of a random intervention $do(A \coloneqq \nu)$, and we select $\lambda$ for $\gamma_{\text{AR}}$ such that this intervention is contained within $C_A(\lambda)$.
We then simulate test data from two distributions:  $\P_{do(A \coloneqq \nu)}$ (i.e., the shift occurs), and $\P$ (where it does not), and evaluate the mean squared prediction error (MSPE). The results are shown in Figure~\ref{fig:targeted-shift}, and demonstrated that TAR performs better than AR and OLS in the first scenario, but this comes at the cost of worse performance on the training distribution.

\paragraph{Technical details}
The entire procedure below produces a prediction MSE for each of three methods and two settings, and we repeat this $m = 10^5$ times, to produce the histograms of MSEs shown in Figure~\ref{fig:targeted-shift}.

We simulate a training data set $\mathcal{D}_{\text{train}}$ ($n_{\text{train}} = 10^5$) from the structural equations
\begin{align*}
    A &:= \epsilon_A \\
    (Y, X, H) &:= (\operatorname{Id}-B)^{-1}(M A + \epsilon),
\end{align*}
where $d_A = d_X = 2$ and $d_Y = d_H$ = 1, $\epsilon_A, \epsilon \sim \mathcal{N}(0, \operatorname{Id})$ and $M$ and $B$ were selected by a simulation resulting in:
\begin{align*}
    M = \begin{pmatrix}
    2& 1\\
    0& 1\\
    2& 2\\
    0& 3
    \end{pmatrix}, 
    B = 
    \begin{pmatrix}
        0  & -0.06&  0.07&  0.04\\
        0.05&  0  &  0.19&  0.03\\
        0.11& -0.11&  0  &  0.1 \\
        -0.02&  0.02&  0.09 &  0 
    \end{pmatrix}.
\end{align*}
We consider the target distribution $do(A:=\kappa^\top A + \eta)$ where 
\begin{align*}
    \kappa = \begin{pmatrix} \sqrt{2} & 0 \\ 0 & 1\end{pmatrix}, 
    \eta = \begin{pmatrix} 0 \\ 2 \end{pmatrix}, 
\end{align*}
and so we fit the targeted AR estimator $(\gamma_{\text{targeted-AR}}, \alpha_{\text{targeted-AR}})$ from Equation~\eqref{eq:tar_a}, where the covariance of the anticipated shift is given by $\Sigma_\nu := \kappa^\top \E[AA^\top]\kappa$, and the mean shift is simply $\eta$.
We also fit OLS estimates $\gamma_{\text{OLS}}(X, Y)$ and $\gamma_{\text{AR}}(X, Y, A)$ where for AR we select $\lambda$ such that $(1+\lambda)$ equals the largest eigenvalue of $\kappa^\top\E[AA^\top]\kappa + \eta\eta^\top$, such that $\E[(\kappa^\top A + \eta)(\kappa^\top A + \eta)^\top] \preceq (1+\lambda)\E[AA^\top]$.

We then simulate a test data set $(n_{\text{test}}=10^5)$ both from 1) the training distribution (i.e. same simulation procedure as for the training set) or 2) by changing the structural equation for $A$ to $A := \kappa^\top \epsilon_A + \eta$, and keeping all other quantities as for the simulation of training data (i.e. the test distribution is the anticipated distribution). We evaluate the prediction MSE on each of the data sets by $\frac{1}{n_{\text{test}}}\sum_j (Y_j - \gamma^\top X_j)^2$ (including the term $\alpha_{\text{targeted-AR}}$ for the targeted AR).

\subsection{Details of Section~\ref{sec:experiment_pollution}}
\label{sub:details_of_section_ref_sec_experiment_pollution}

\paragraph{Features} The dataset contains time-stamps as well as season indicators, which we do not use anywhere as features.  The remaining features are Dew Point (Celsius Degree), Temperature (Celsius Degree), Humidity (\%), Pressure (hPa), Combined wind direction (NE, NW, SE, SW, or CV, indicating calm and variable), Cumulated wind speed (m/s), Hourly precipitation (mm), and Cumulated precipitation (mm).

\paragraph{Data Processing} Each city has PM2.5 readings from multiple sites, which we average to get a single reading, and we take a log transformation.  For Precipitation (Cumulative) we subtract off the (current hour) precipitation to avoid co-linearity. We take a log transformation of the variable for Wind Speed, Precipitation (Hourly) and Precipitation (Cumulative), due to skewness.  We drop all rows that contain any missing data.

\paragraph{Proxies (Temperature)} We use temperature as our proxy variable, and treat it as unavailable at test time.  We construct two synthetic proxies of temperature to serve as $W, Z$, adding independent Gaussian noise while controlling the signal-to-variance ratio (in the training distribution) at $\var(A) / \var(W) = 0.9$.  This results in different standard deviations of the Gaussian noise across different environments, because of differences in the training distributions across training seasons and cities.  The standard error of the noise varies between 2 and 5 degrees, to maintain the same signal-to-variance ratio.

\paragraph{Training Details (PAR, xPAR)} For the distributional robustness approaches described in Section~\ref{sec:worst_case_bounded_shift}, we choose $\lambda \in [0, 40]$ by leave-one-group-out cross-validation on the three training seasons, using the first year (2013) of data.  For Proxy Anchor Regression using Temperature directly, there is heterogeneity in the cross-validated choice of $\lambda$: In 9 out of 20 scenarios, $\lambda = 40$ is chosen, but in the remaining 11, $\lambda = 0$ is chosen, which is equivalent to OLS\@.  We saw a similar result when the maximum value of $\lambda$ was 20, and increased the maximum limit to 40 without seeing much difference, so we did not increase it further. 
Concretely, with $\lambda$ in [0, 20], there are some scenarios where PAR (TempC) has slightly worse or slightly better MSE (vs. $\lambda$ in [0, 40]), but the differences are all less than 0.001. The only observable difference in Table \ref{tab:mse_pollution} when running with $\lambda$ in [0, 20] is that the “best” performance is -0.040 ($\lambda=20$), as opposed to -0.041 ($\lambda = 40$) [where lower is better, rounded to nearest 0.001].  For Proxy Anchor Regression using $W$ and for Cross-Proxy Anchor Regression (xPAR) using $W, Z$ together, we use the same values of $\lambda$ as above, for comparability.

\paragraph{Training Details (PTAR, xPTAR)}For the targeted approaches described in Section~\ref{sec:targeted_shift}, we use the mean and variance of the temperature in the test distribution to target our predictors, and similarly use the distribution of the proxies when using Proxy Targeted Anchor Regression (PTAR) with $W$ and Cross-Proxy TAR (xPTAR) with $W, Z$.  Note that xPTAR (unlike xPAR) is asymmetric in the proxies, but in this case the proxies are distributed identically.

\paragraph{Benchmarks} As described in the main text, our primary benchmark is OLS, trained on the three training seasons, evaluated on the held-out season.  We also include two other baselines:  First, OLS that has access to temperature during both train and test, which we denote OLS (TempC), and OLS that includes temperature during training, and attempts to estimate a bias term by plugging in the mean (test) value for temperature during prediction.  

In Table~\ref{tab:mse_pollution_appendix} we give the full results over all 20 scenarios, which includes the 11 scenarios where $\lambda = 0$ is chosen by cross-validation, rendering the PAR and xPAR solutions equivalent to OLS\@.

\begin{table}[t]
  \centering
  \caption{MSE (lower is better) over 20 scenarios consisting of five cities and four held-out seasons. Average difference to OLS estimator (lower is better) given in the second column, and minimum / maximum difference in remaining columns.}%
  \label{tab:mse_pollution_appendix}
  \begin{tabular}{lrrrr}
  \toprule
  Estimator &  Mean &   Diff &    Min &   Max \\
  \midrule
  OLS               & 0.457 &   &   &  \\
  OLS (TempC)       & 0.455 & -0.002 & -0.028 & 0.026 \\
  OLS + Est. Bias   & 0.474 &  0.018 & -0.072 & 0.150 \\
  \midrule
  PAR (TempC)        & 0.454 & -0.003 & -0.041 & 0.006 \\
  PAR (W)            & 0.454 & -0.002 & -0.037 & 0.006 \\
  xPAR (W, Z)        & 0.454 & -0.003 & -0.039 & 0.007 \\
  \midrule
  PTAR               & 0.450 & -0.007 & -0.061 & 0.002 \\
  PTAR (W)           & 0.452 & -0.005 & -0.038 & 0.001 \\
  xPTAR (W, Z)       & 0.450 & -0.007 & -0.059 & 0.003 \\
  \bottomrule
  \end{tabular}
\end{table}

\begin{figure}[t]
\centering
  \includegraphics[width=\linewidth]{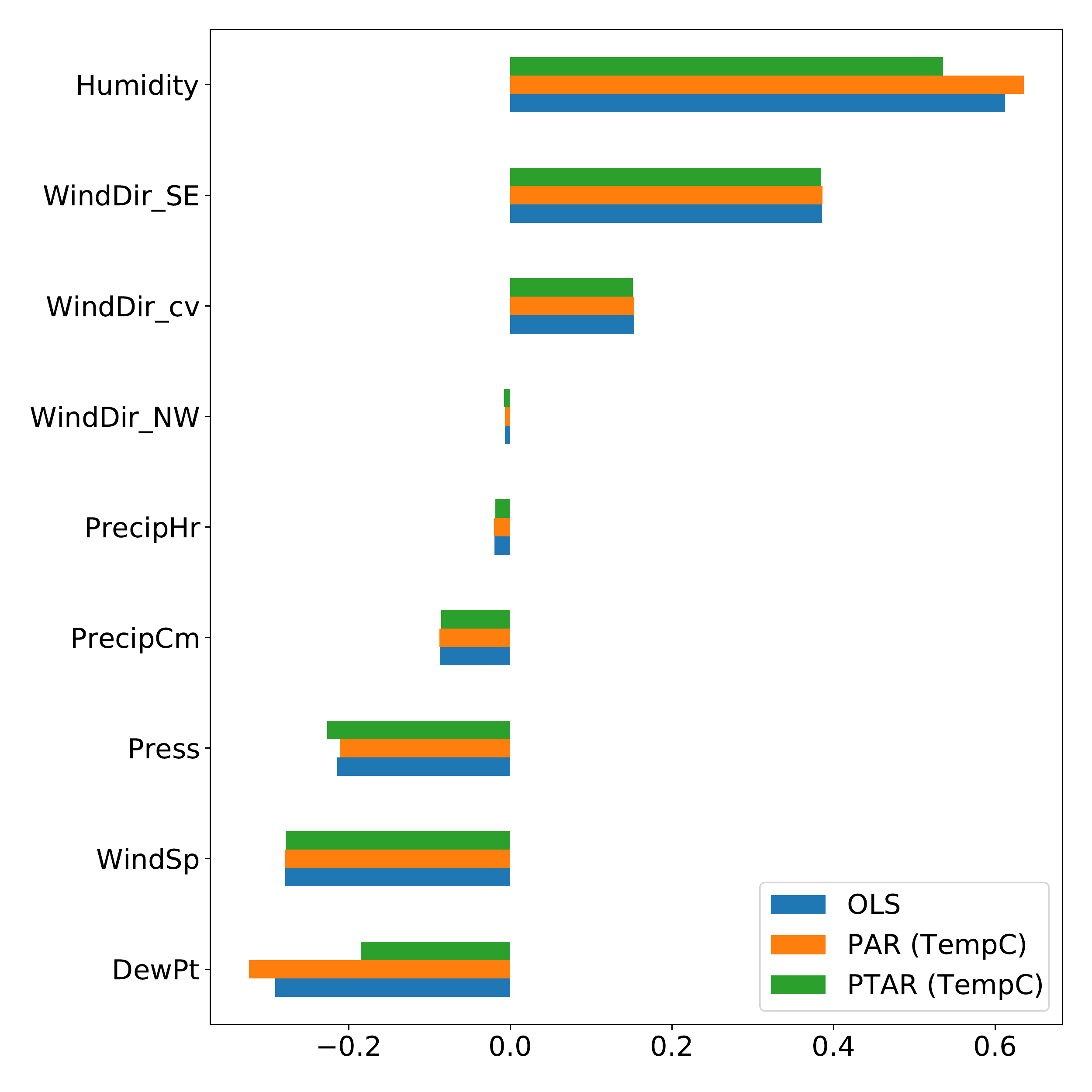}
  \caption{Comparison of learned coefficients.  All variables were standardized to unit variance. The intercept for OLS and AR is the same (by construction) at $\alpha = 4.087$ while the intercept for TAR is lower at $\alpha = 3.885$.}\label{fig:comparison_coefficients}
\end{figure}

\begin{figure}[t]
    \centering
    \includegraphics[width=\linewidth]{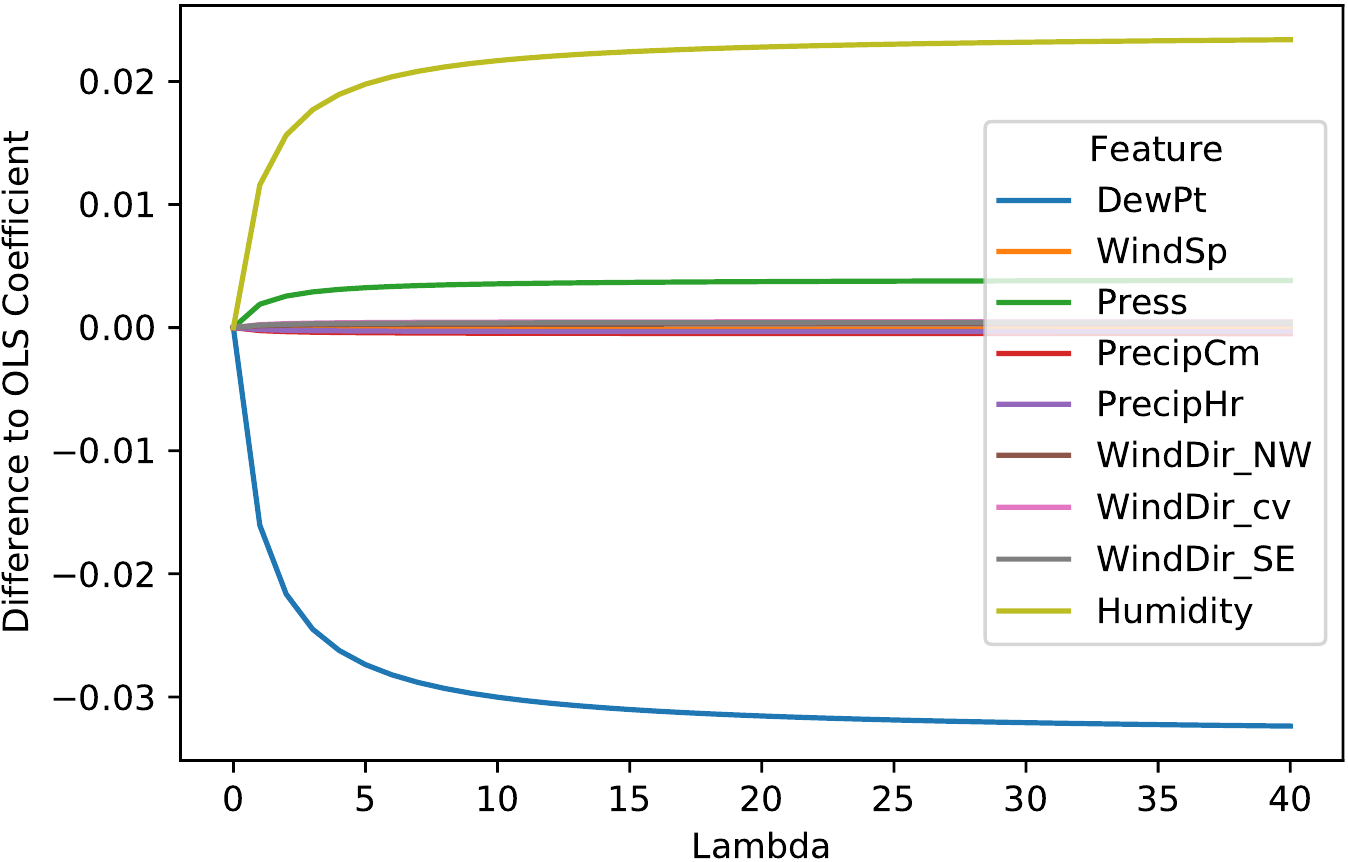}
    \caption{Coefficient path, showing the difference between the PAR and OLS coefficients in Figure~\ref{fig:comparison_coefficients} for different values of $\lambda$.}
    \label{fig:coefficient_path}
\end{figure}

\paragraph{Regularization paths} In Figure~\ref{fig:comparison_coefficients} we have shown how the solution in the \enquote{best} scenario differs for Proxy Anchor Regression (PAR) with $\lambda = 40$ versus OLS (i.e., $\lambda = 0$).  In Figure~\ref{fig:coefficient_path}, we show how the coefficients change in-between these two extremes: for every integer value of $\lambda$ in [0, 40] we show the difference in the PAR vs. OLS coefficients for each feature. Increasing $\lambda$ further does not make a significant difference for this particular example.

\begin{figure}[t]
\centering
  \includegraphics[width=\linewidth]{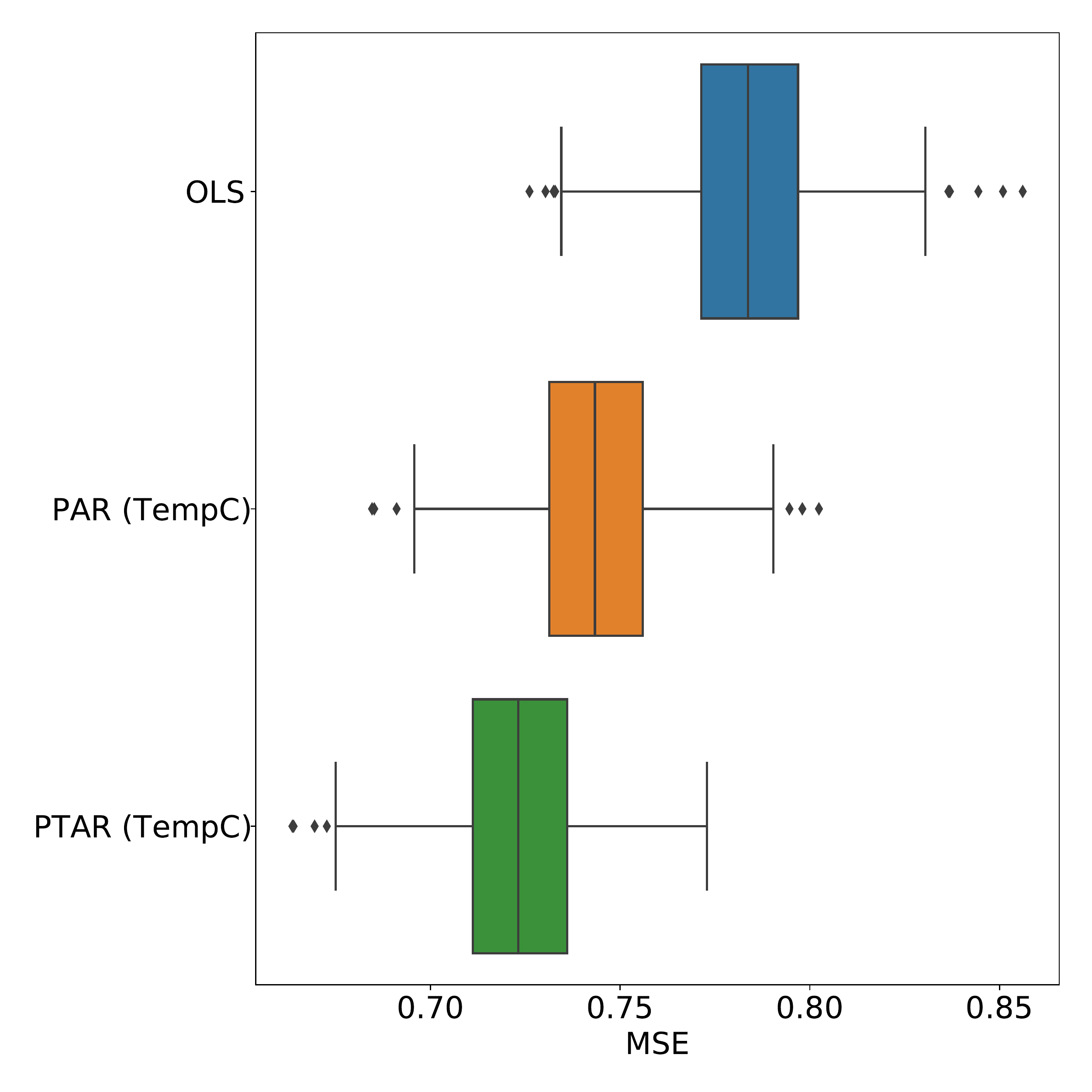}
  \caption{Best performance for Proxy Anchor Regression (PAR) and Proxy Targeted AR (PTAR), corresponding to Summer in Beijing.  Variance estimates generated by bootstrapping the test residuals of the fitted models.}
  \label{fig:bootstrap_best_case}
\end{figure}

\FloatBarrier

\section{Additional experiment: Signal-to-variance ratio}%
\label{sec:addl_exp_svr}

\begin{figure}[t]
    \centering
    \input{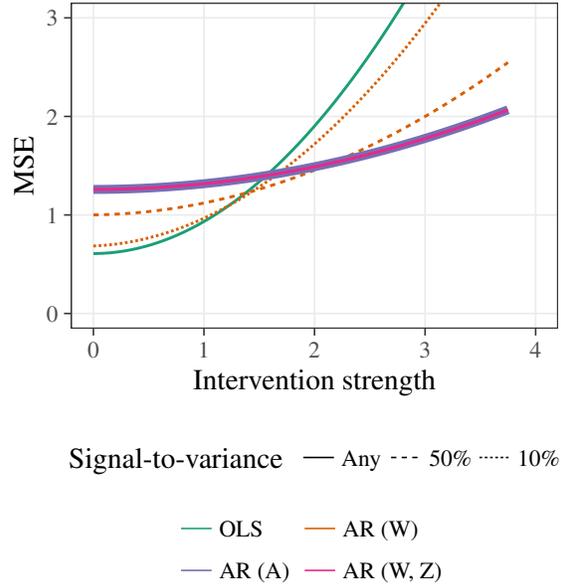}
    \vspace{-10pt}
    \caption{Anchor and proxy estimators for different levels of signal-to-variance ratio $\beta(\E[WW^\top])^{-1}\beta^\top$. A training data set ($n = 10^7$) with two proxies $W, Z$ is simulated and the estimators $\hat\gamma_{\text{PAR}(A)}$,  $\hat\gamma_{\text{xPAR}(W, Z)}$, $\hat\gamma_{\text{AR}(A)}$, and $\hat\gamma_{\text{OLS}}$ are fitted using a fixed $\lambda$.
    Interventions $v$ of increasing strength is sampled, and for each a new data set ($n = 10^5$) is sampled from $\P^{do(A:=v)}$, and for each estimator $\hat\gamma$, the prediction mean squared error $\E_{do(A:=(v_1, v_2))}[(Y - \hat\gamma^\top X)^2]$ is computed. 
    This procedure is repeated for signal-to-variance ratios $10\%$ and $50\%$.
    }  
    \label{fig:proxy-intervention}
\end{figure}
To examine the effect of the signal strengths $\beta_W$ and $\beta_Z$, we scale the signals $\beta_{W,s} = \beta_{Z,s} = s\operatorname{Id}$ for $s\in \{0, \sqrt{2/3}, 0.8\}$, which for the single proxy estimator $\hat\gamma_{\text{PAR}}$ amounts to optimizing over worst case loss in the robustness regions $C(\lambda) = \{vv\top \preceq (1 + \lambda \frac{s^2}{1+s^2})\operatorname{Id}\}$. 

For $s \in \{1, 3\}$, such that the signal-to-variance ratio $\frac{s^2}{1+s^2}$ equals either $10\%$ or $50\%$, we simulate a training data set $\mathcal{D}_{\text{train}}$ with two proxies $W$ and $Z$ from the structural equations $A := \epsilon_A, (X^\top, Y^\top, H^\top)^\top := (1-B)^{-1}(M_a A + \epsilon), W:= \beta_{W,s}^\top A + \epsilon_W$ and $Z := \beta_{Z,s}^\top A + \epsilon_Z$ where all noise terms are i.i.d\ with unit covariance and $M_A, B$ are given by:
\begin{align*}
    M := \begin{pmatrix}
    2 & 1 \\ 
    0 & 1 \\
    2 & 2 \\
    0 & 3
    \end{pmatrix}, B := 
    \begin{pmatrix}
    0  & -0.57 &  0.73&  0.37 \\
    0.53&  0 & 1.91 & 0.33 \\
    1.14& -1.13 &  0  &  0.96 \\
    -0.22&  0.16 &  0.87 &  0
    \end{pmatrix}.
\end{align*}
Since for this experiment we are not interested in finite sample properties of the estimators, we use sample size $n = 10^7$.

For each data set we fit estimators $\hat\gamma_{\text{PAR}(W)}$ (using only one proxy), $\hat\gamma_{\text{xPAR(W, Z)}}$ (using both proxies), $\hat\gamma_{\text{AR}(A)}$, and $\hat\gamma_{\text{OLS}}$, and evaluate the estimators at data sampled from interventional distributions $\P_{do(A:=v)}$ for several interventions $v$ of increasing strength (i.e. increasing distance from $\E[A] = 0$). 

As the signal to variance ratio increases, the PAR($W$) loss approaches the AR($A$). Further we observe that xPAR($W, Z$) coincides with the $AR(A)$ estimator for both signal-to-variance levels.  This is illustrated in Figure~\ref{fig:proxy-intervention}.

\clearpage
\bibliography{main.bbl}

\begin{thebibliography}{32}
\providecommand{\natexlab}[1]{#1}
\providecommand{\url}[1]{\texttt{#1}}
\expandafter\ifx\csname urlstyle\endcsname\relax
  \providecommand{\doi}[1]{doi: #1}\else
  \providecommand{\doi}{doi: \begingroup \urlstyle{rm}\Url}\fi

\bibitem[Arjovsky et~al.(2019)Arjovsky, Bottou, Gulrajani, and
  Lopez-Paz]{Arjovsky2019-kv}
Arjovsky, M., Bottou, L., Gulrajani, I., and Lopez-Paz, D.
\newblock Invariant risk minimization.
\newblock \emph{arXiv (1907.02893)}, 2019.

\bibitem[Bellot \& van~der Schaar(2020)Bellot and van~der
  Schaar]{Bellot2020-dd}
Bellot, A. and van~der Schaar, M.
\newblock {Accounting for Unobserved Confounding in Domain Generalization}.
\newblock \emph{arXiv (2007.10653)}, July 2020.

\bibitem[Bound et~al.(2001)Bound, Brown, and Mathiowetz]{Bound2001}
Bound, J., Brown, C., and Mathiowetz, N.
\newblock {Chapter 59: Measurement Error In Survey Data}.
\newblock \emph{Handbook of Econometrics}, 5:\penalty0 3705--3843, 2001.

\bibitem[B{\"{u}}hlmann \& {\'{C}}evid(2020)B{\"{u}}hlmann and
  {\'{C}}evid]{Buhlmann2020b}
B{\"{u}}hlmann, P. and {\'{C}}evid, D.
\newblock Deconfounding and causal regularisation for stability and external
  validity.
\newblock \emph{International Statistical Review}, 88\penalty0 (S1):\penalty0
  S114--S134, 2020.

\bibitem[Christiansen et~al.(2020)Christiansen, Pfister, Jakobsen, Gnecco, and
  Peters]{Christiansen2020}
Christiansen, R., Pfister, N., Jakobsen, M.~E., Gnecco, N., and Peters, J.
\newblock {The Difficult Task of Distribution Generalization in Nonlinear
  Models}.
\newblock \emph{arXiv}, pp.\  1--48, 2020.
\newblock URL \url{http://arxiv.org/abs/2006.07433}.

\bibitem[Duchi et~al.(2020)Duchi, Hashimoto, and Namkoong]{Duchi2020}
Duchi, J.~C., Hashimoto, T., and Namkoong, H.
\newblock {Distributionally robust losses for latent covariate mixtures}.
\newblock \emph{arXiv (2007.13982)}, pp.\  1--39, 2020.

\bibitem[Frost \& Thompson(2000)Frost and Thompson]{Frost2000}
Frost, C. and Thompson, S.~G.
\newblock {Correcting for Regression Dilution Bias: Comparison of Methods for a
  Single Predictor Variable}.
\newblock \emph{Journal of the Royal Statistical Society: Series A},
  163\penalty0 (2):\penalty0 173--189, 2000.

\bibitem[Fuller(1987)]{Fuller1987}
Fuller, W.~A.
\newblock \emph{{Measurement error models}}.
\newblock John Wiley and Sons Inc., 1987.

\bibitem[Guo et~al.(2021)Guo, Zhang, Liu, and Kiciman]{Guo2021-bn}
Guo, R., Zhang, P., Liu, H., and Kiciman, E.
\newblock {Out-of-distribution Prediction with Invariant Risk Minimization: The
  Limitation and An Effective Fix}.
\newblock \emph{arXiv (2101.07732)}, January 2021.

\bibitem[Hern{\'{a}}n \& Robins(2006)Hern{\'{a}}n and Robins]{Hernan2006}
Hern{\'{a}}n, M.~A. and Robins, J.~M.
\newblock {Instruments for causal inference: An epidemiologist's dream?}
\newblock \emph{Epidemiology}, 17\penalty0 (4):\penalty0 360--372, 2006.

\bibitem[Hyslop \& Imbens(2001)Hyslop and Imbens]{Hyslop2001}
Hyslop, D.~R. and Imbens, G.~W.
\newblock {Bias from classical and other forms of measurement error}.
\newblock \emph{Journal of Business and Economic Statistics}, 19\penalty0
  (4):\penalty0 475--481, 2001.

\bibitem[Jakobsen \& Peters(2020)Jakobsen and Peters]{Jakobsen2020}
Jakobsen, M.~E. and Peters, J.
\newblock {Distributional Robustness of K-class Estimators and the PULSE}.
\newblock \emph{arXiv (2005.03353)}, 2020.

\bibitem[Kamath et~al.(2021)Kamath, Tangella, Sutherland, and
  Srebro]{Kamath2021-ww}
Kamath, P., Tangella, A., Sutherland, D.~J., and Srebro, N.
\newblock {Does Invariant Risk Minimization Capture Invariance?}
\newblock \emph{arXiv (2101.01134)}, January 2021.

\bibitem[Kook et~al.(2021)Kook, Sick, and B{\"u}hlmann]{Kook2021-lk}
Kook, L., Sick, B., and B{\"u}hlmann, P.
\newblock {Distributional Anchor Regression}.
\newblock \emph{arXiv (2101.08224)}, January 2021.

\bibitem[Krueger et~al.(2020)Krueger, Caballero, Jacobsen, Zhang, Binas, Zhang,
  Le~Priol, and Courville]{Krueger2020-np}
Krueger, D., Caballero, E., Jacobsen, J.-H., Zhang, A., Binas, J., Zhang, D.,
  Le~Priol, R., and Courville, A.
\newblock {Out-of-Distribution Generalization via Risk Extrapolation (REx)}.
\newblock \emph{arXiv (2003.00688)}, March 2020.

\bibitem[Kuroki \& Pearl(2014)Kuroki and Pearl]{Kuroki2014}
Kuroki, M. and Pearl, J.
\newblock {Measurement bias and effect restoration in causal inference}.
\newblock \emph{Biometrika}, 101\penalty0 (2):\penalty0 423--437, 2014.

\bibitem[Liang et~al.(2016)Liang, Li, Zhang, Huang, and Chen]{Liang2016}
Liang, X., Li, S., Zhang, S., Huang, H., and Chen, S.~X.
\newblock {PM2.5 data reliability, consistency, and air quality assessment in
  five Chinese cities}.
\newblock \emph{Journal of Geophysical Research: Atmospheres}, 121, 2016.

\bibitem[Lovell(1963)]{Lovell1963}
Lovell, M.~C.
\newblock Seasonal adjustment of economic time series and multiple regression
  analysis.
\newblock \emph{Journal of the American Statistical Association}, 58\penalty0
  (304):\penalty0 993--1010, 1963.

\bibitem[Lovell(2008)]{Lovell2008}
Lovell, M.~C.
\newblock {A simple proof of the FWL theorem}.
\newblock \emph{Journal of Economic Education}, 39\penalty0 (1):\penalty0
  88--91, 2008.

\bibitem[Magliacane et~al.(2018)Magliacane, van Ommen, Claassen, Bongers,
  Versteeg, and Mooij]{Magliacane2017}
Magliacane, S., van Ommen, T., Claassen, T., Bongers, S., Versteeg, P., and
  Mooij, J.~M.
\newblock Domain adaptation by using causal inference to predict invariant
  conditional distributions.
\newblock In \emph{Proceedings of the 32nd Conference on Neural Information
  Processing Systems (NeurIPS)}, 2018.

\bibitem[Miao \& Tchetgen(2018)Miao and Tchetgen]{Miao2018bridge}
Miao, W. and Tchetgen, E.~T.
\newblock {A Confounding Bridge Approach for Double Negative Control Inference
  on Causal Effects}.
\newblock \emph{arXiv (1808.04945)}, 2018.

\bibitem[Pearl(2009)]{Pearl2009}
Pearl, J.
\newblock \emph{{Causality: Models, Reasoning, and Inference}}.
\newblock Cambridge University Press, 2nd edition, 2009.

\bibitem[Qui{\~n}onero-Candela et~al.(2009)Qui{\~n}onero-Candela, Sugiyama,
  Schwaighofer, and Lawrence]{Quinonero-Candela2009}
Qui{\~n}onero-Candela, J., Sugiyama, M., Schwaighofer, A., and Lawrence, N.~D.
  (eds.).
\newblock \emph{{Dataset Shift in Machine Learning}}.
\newblock MIT Press, 2009.

\bibitem[Rojas-Carulla et~al.(2018)Rojas-Carulla, Sch{\"{o}}lkopf, Turner, and
  Peters]{Rojas-Carulla2015}
Rojas-Carulla, M., Sch{\"{o}}lkopf, B., Turner, R., and Peters, J.
\newblock {Invariant Models for Causal Transfer Learning}.
\newblock \emph{Journal of Machine Learning Research}, 19\penalty0
  (36):\penalty0 1--34, 2018.

\bibitem[Rosenfeld \& Risteski(2020)Rosenfeld and Risteski]{Rosenfeld2020}
Rosenfeld, E. and Risteski, A.
\newblock {The Risks of Invariant Risk Minimization}.
\newblock \emph{arXiv (2010.05761v1)}, pp.\  1--36, 2020.

\bibitem[Rothenh{\"{a}}usler et~al.(2021)Rothenh{\"{a}}usler, Meinshausen,
  B{\"{u}}hlmann, and Peters]{Rothenhausler2018}
Rothenh{\"{a}}usler, D., Meinshausen, N., B{\"{u}}hlmann, P., and Peters, J.
\newblock Anchor regression: Heterogeneous data meet causality.
\newblock \emph{Journal of the Royal Statistical Society: Series B (Statistical
  Methodology)}, 83\penalty0 (2):\penalty0 215--246, 2021.

\bibitem[Shi et~al.(2018)Shi, Miao, Nelson, and Tchetgen]{Shi2018}
Shi, X., Miao, W., Nelson, J.~C., and Tchetgen, E. J.~T.
\newblock {Multiply Robust Causal Inference with Double Negative Control
  Adjustment for Categorical Unmeasured Confounding}.
\newblock \emph{arXiv (1808.04906)}, 2018.

\bibitem[Srivastava et~al.(2020)Srivastava, Hashimoto, and
  Liang]{Srivastava2020}
Srivastava, M., Hashimoto, T., and Liang, P.
\newblock {Robustness to Spurious Correlations via Human Annotations}.
\newblock \emph{37th International Conference on Machine Learning}, 2020.

\bibitem[Subbaswamy et~al.(2019)Subbaswamy, Schulam, and Saria]{Subbaswamy2019}
Subbaswamy, A., Schulam, P., and Saria, S.
\newblock {Preventing Failures Due to Dataset Shift: Learning Predictive Models
  That Transport}.
\newblock In \emph{Proceedings of the 22nd International Conference on
  Artificial Intelligence and Statistics (AISTATS)}, 2019.

\bibitem[{Tchetgen Tchetgen} et~al.(2020){Tchetgen Tchetgen}, Ying, Cui, Shi,
  and Miao]{TchetgenTchetgen2020}
{Tchetgen Tchetgen}, E.~J., Ying, A., Cui, Y., Shi, X., and Miao, W.
\newblock {An Introduction to Proximal Causal Learning}.
\newblock \emph{arXiv (2009.10982)}, 2020.

\bibitem[Xie et~al.(2020)Xie, Ye, Chen, Liu, Sun, and Li]{Xie2020-fn}
Xie, C., Ye, H., Chen, F., Liu, Y., Sun, R., and Li, Z.
\newblock {Risk Variance Penalization}.
\newblock \emph{arXiv (2006.07544)}, June 2020.

\bibitem[Zhang(2006)]{zhang2006schur}
Zhang, F.
\newblock \emph{The Schur complement and its applications}, volume~4.
\newblock Springer Science \& Business Media, 2006.

\end{thebibliography}
\bibliographystyle{icml2021}

\end{document}